\documentclass[twoside,11pt]{article}

\usepackage{jmlr2e}

\usepackage{amsfonts}
\usepackage{nicefrac}

\newcommand{\mI}[0]{\textnormal{I}}

\newcommand{\mJ}[0]{\textnormal{J}}
\newcommand{\mA}[0]{\textnormal{A}}

\newcommand{\mH}[0]{\textnormal{H}}

\newcommand{\BR}{{\mathbb{R}}}
\newcommand{\BE}{{\mathbb{E}}}

\newcommand{\CB}{{\mathcal{B}}}

\newcommand{\CN}{\mathcal{N}}
\newcommand{\CO}{{\mathcal{O}}}

\newcommand{\data}[0]{{y}}

\newcommand{\op}[0]{\mA}
\newcommand{\params}[0]{{{\theta}}}
\newcommand{\TV}[0]{{\rm TV}}

\providecommand{\argmin}{\operatorname*{argmin}}

\newcommand{\logdet}[0]{{\rm logdet}}

\usepackage{amsmath}
\usepackage{xcolor}

\usepackage{nth}
\usepackage{natbib}
\usepackage{nicematrix}
\usepackage{arydshln}
\usepackage{tikz}
\NiceMatrixOptions{letter-for-dotted-lines=V}
\usepackage{tabularx}
\usepackage{leftidx,mathtools}
\usepackage{enumitem}
\usepackage[ruled, vlined, linesnumbered,algo2e]{algorithm2e}

\usepackage{cleveref}

\SetKwInput{KwData}{Inputs}
\SetKwInput{KwResult}{Output}

\jmlrheading{1}{}{}{}{}{}
\firstpageno{1}

\makeatletter
\newcommand{\printfnsymbol}[1]{%
  \textsuperscript{\@fnsymbol{#1}}%
}
\makeatother

\begin{document}

\title{Bayesian Experimental Design for Computed Tomography with the Linearised Deep Image Prior}

\author{\name Riccardo Barbano\thanks{Authors contributed equally. Our code is at \href{https://github.com/educating-dip/bayesian_experimental_design}{\tt github.com/educating-dip/bayesian\_experimental\_design}} \email riccardo.barbano.19@ucl.ac.uk \\
\addr Department of Computer Science, University College London
\AND
\name Johannes Leuschner\printfnsymbol{1} \email jleuschn@uni-bremen.de \\
\addr Center for Industrial Mathematics, University of Bremen
\AND
\name Javier Antorán\printfnsymbol{1} \email ja666@cam.ac.uk \\
\addr Department of Engineering, University of Cambridge
\AND
\name Bangti Jin \email b.jin@ucl.ac.uk \\
\addr Department of Computer Science, University College London
\AND
\name José Miguel Hernández-Lobato \email jmh233@cam.ac.uk \\
\addr Department of Engineering, University of Cambridge
}

\maketitle

\begin{abstract}%
We investigate adaptive design based on a single sparse pilot scan for generating effective scanning strategies for computed tomography reconstruction. We propose a novel approach using the linearised deep image prior. It allows incorporating information from the pilot measurements into the angle selection criteria, while maintaining the tractability of a conjugate Gaussian-linear model. On a synthetically generated dataset with preferential directions, linearised DIP design allows reducing the number of scans by up to 30\% relative to an equidistant angle~baseline.
\end{abstract}

\section{Introduction and related work}\label{sec:intro}

Linear inverse problems in imaging aim to recover an unknown image ${x} \in \BR^{d_{x}}$ from measurements $\data \in \BR^{d_{y}}$, which are often described by the application of a forward operator $\op \in \BR^{d_{y} \times d_{x}}$, and the addition of Gaussian noise $\epsilon \sim \CN(0, \sigma^{2}_{y} \, \mI_{d_{y}})$ as
\begin{equation}\label{eqn:inverse_problem}
    \data = \op {x} + \epsilon.
\end{equation}
This acquisition model is ubiquitous in machine vision, computed tomography (CT), and magnetic resonance imaging among other applications.
Due to the inherent ill-posedness of the task (e.g.\ $d_{y} \ll d_{x}$), suitable regularisation or prior assumptions are crucial for the stable and accurate recovery of $x$ \citep{tikhonov1977solutions, ito2014inverse}.
In this work, we focus on X-ray imaging, a setting with application to both medical and industrial settings  \citep{buzug2011computed}.

\begin{figure}[t]
    \vspace{-1.05cm}
    \begin{minipage}{.3\textwidth}
        \centering
        \includegraphics[width=\linewidth]{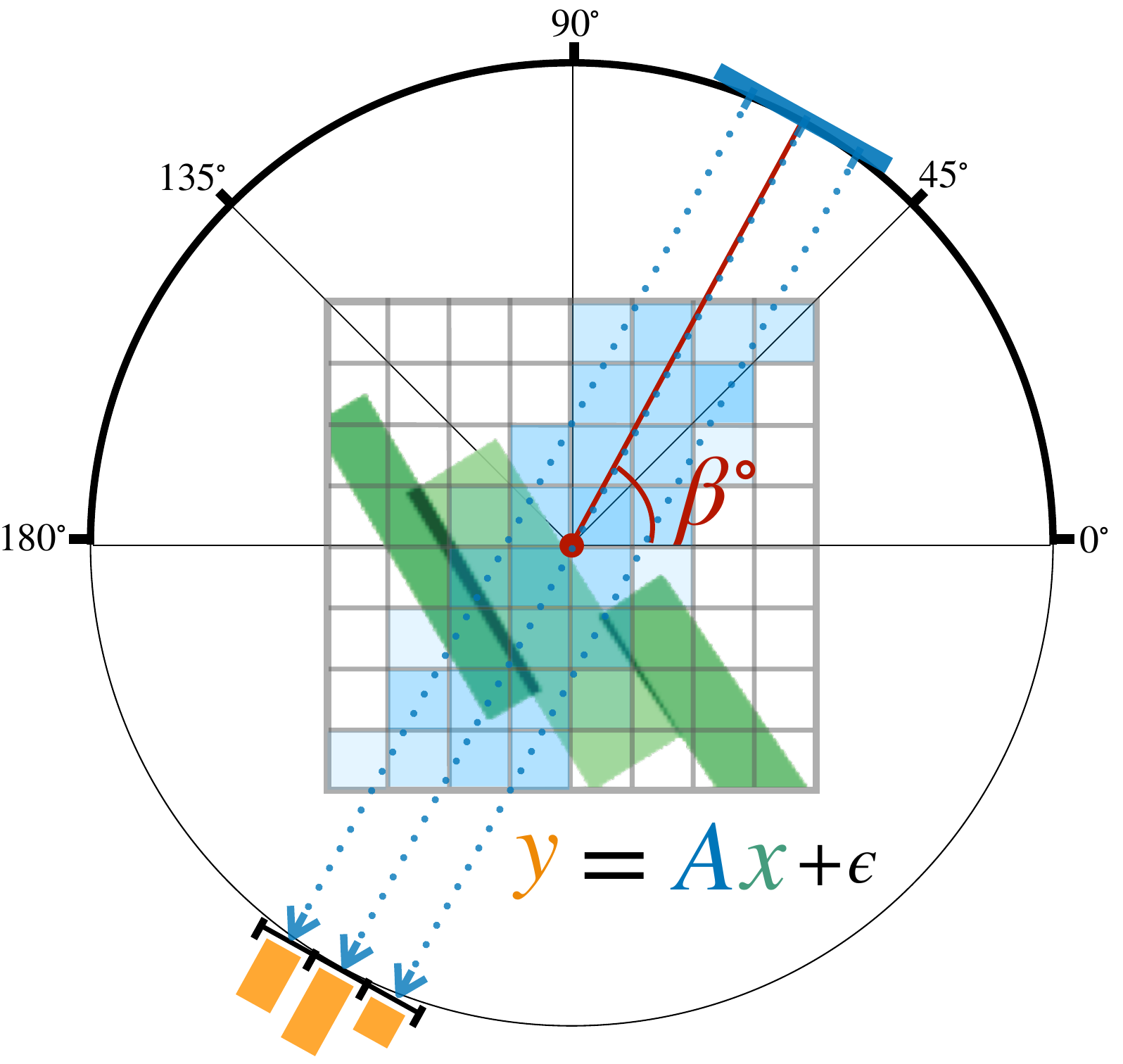}
    \end{minipage}%
    \begin{minipage}{0.7\textwidth}
        \centering
        \includegraphics[width=\linewidth]{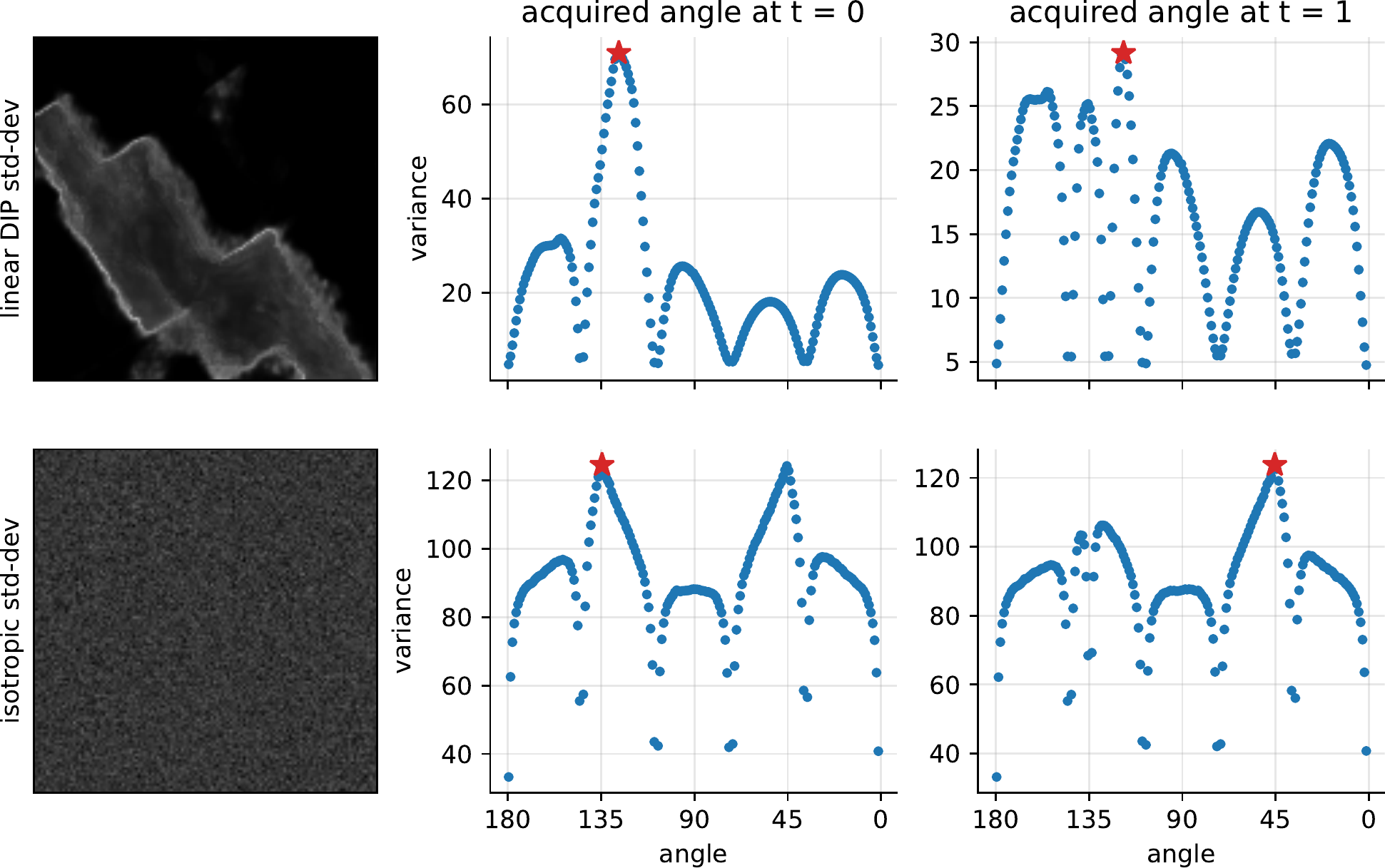}
    \end{minipage}
    \vspace{-0.4cm}
    \caption{Left: A schematic diagram of 2D parallel beam CT geometry, used in the experiments. Top row: the linearised DIP assigns prior variance to pixels where edges are present, guiding angle selection so that X-ray quanta cover these pixels. Bottom row: the isotropic linear model's variance does not depend on the measurements. Angles $45,135$ are chosen since they are oblique and maximise quanta path-length in the image.}
    \vspace{-0.8cm}
    \label{fig:variance_angles}
\end{figure}

In CT, an emitter sends X-ray quanta through the object being scanned. The quanta are captured by $d_{p}$ detector elements placed opposite the emitter. Each row of $\op$ tells us about which regions (pixels) the X-ray quanta will pass through before reaching a detector element (cf.\ \cref{fig:variance_angles}). The number of X-ray quanta measured by a detector pixel conveys information about the attenuation coefficient of the material present along the quanta's path.
This procedure is repeated at $d_{\CB}$ angles, yielding a measurement of dimension $d_{y}=d_{p}\cdot d_{\CB}$.

In CT, Bayesian experimental design employs prior assumptions to select scanning angles which are aimed to yield the highest fidelity reconstruction. 
Adaptive design further incorporates information gained at previous angles to inform subsequent angle selections \citep{Chaloner1995review}. 
These methods are of great practical interest since they promise to reduce radiation dosages and scanning times.
Alas, existing CT design methods often struggle to improve over equidistant angle choice \citep{Shen2022learningtoscan}. 
Furthermore, the requisite of additional computations before subsequent scans makes adaptive methods impractical for many~applications.

Critically important to experimental design is the choice of prior \citep{chi2015modelerror,foster2021Design}.
Linear models allow for tractable computation of quantities of interest for design, but their predictive uncertainty is independent of previously measured values, disallowing adaptive design \citep{Burger2021lineardesign}. 
More complex model choices make inference difficult, necessitating approximations which can degrade performance \citep{Helin2021GaussianTV,Shen2022learningtoscan}.

This work aims to make adaptive design practical by considering a setting where the CT scan is performed in two phases. 
First, a sparse pilot scan is performed to provide data with which to fit adaptive methods.
These are then used to select angles for a full scan. 
We demonstrate this procedure with a synthetic dataset where a different ``preferential'' angle is most informative for each image.
Preferential directions appear commonly in industrial CT for material science and in medical CT for medical implant assessment.
We use the linearised Deep Image Prior (DIP) \citep{barbano2022probabilistic} as a data-dependent prior for adaptive design which preserves the tractability of conjugate Gaussian-linear models.
Unlike simple linear models, the linearised DIP outperforms the equidistant angle baseline.
Finally, we show that designs obtained with the linearised DIP perform well under traditional (non DIP-based) regularised-reconstruction.

\section{Regularised reconstruction and deep image prior}\label{sec:preliminaries}

Total Variation (TV) is the most popular regulariser for CT reconstruction \citep{rudin1992nonlinear,chambolle2010introduction}. The anisotropic TV semi-norm of an image vector $x\in\mathbb{R}^{d_x}$ is given by

\begin{align}\label{eq:TV_equation}
    \vspace{-0.5cm}
    \text{TV}(x) = \sum_{i,j} | X_{i,j} - X_{i+1,j}| + \sum_{i,j} |X_{i,j} - X_{i,j+1}|,
\end{align}
where  $X \in \BR^{h \times w}$ denotes the vector $x\in\mathbb{R}^{d_x}$ reshaped into an image of height $h$ by width $w$, and $d_{x} = h\cdot w$.
The corresponding regularised reconstruction is obtained by 
\begin{equation}\label{eq:simple_optimisation_objective}
x^\star\in\argmin_{x\in\mathbb{R}^{d_x}} \|\op x - \data\|^2 + \lambda\mathrm{TV}(x),
\end{equation}
where the hyperparameter $\lambda>0$ determines the strength of regularisation.

The DIP \citep{ulyanov2020dip,baguer2020diptv} reparametrises the reconstruction $x$ as the output of a U-net $x(\theta)$ \citep{ronneberger2015u} with a fixed input, which we omit for clarity, and parameters $\theta \in \BR^{d_{\theta}}$.
The resulting reconstruction problem reads
\begin{equation}\label{eq:DIP_MAP-obj}
    \params^\star\in\argmin_{\params\in\mathbb{R}^{d_\theta}} \|\op x(\theta) - \data\|^2 + \lambda \TV(x(\theta)) \quad \text{and} \quad x^{\star} = x(\theta^{\star}).
\end{equation}
We follow \citet{barbano2021deep} in accelerating optimisation of \cref{eq:DIP_MAP-obj} using pre-trained U-nets.

\section{Linear(ised) models for CT experimental design}\label{sec:methods}

Let $\CB_{a}$ be the set of \emph{all} possible angles at which we can scan. The task is to choose the subset of angles $\CB \subset \CB_{a}$ which produces the highest-fidelity reconstruction. We shall add angles sequentially over $T$ steps. The set $\CB^{(t)}$ denotes the chosen angles up to step $t < T$, and $\bar{\CB}^{(t)}=\CB_{a} \setminus \CB^{(t)}$ the angles left to choose from. $\CB^{(0)}$ denotes the set of angles used in the initial pilot scan, and $\CB = \CB^{(T)}$ the full design.
We incorporate a decision to scan at angle $\beta \in \bar{\CB}^{(t)}$ by concatenating the matrix $\op^{\beta} \in \BR^{d_{p} \times d_{x}}$, which contains a row for each detector pixel at angle $\beta$, to the operator.
After step $t$, the operator $\op^{(t)} \in \BR^{d_{p} \cdot d_{\CB^{(t)}} \times d_{x}}$ stacks $d_{\CB^{(t)}}$ of these matrices, with $d_{\CB^{(t)}}=|\CB^{(t)}|$.
$\bar{\op}^{(t)} \in \BR^{d_{p}\cdot d_{\bar{\CB}^{(t)}} \times d_{x}}$ denotes the forward operator for the angles left to choose~from.

For design, we place a multivariate Gaussian prior on $x$ with zero mean and covariance matrix $\Sigma_{xx} \in \BR^{d_{x} \times d_{x}}$. Together with the Gaussian noise model in \cref{eqn:inverse_problem}, this gives a conjugate Gaussian-linear model. The vector $y^{(t)}\in \BR^{d_{p}\cdot d_{\CB^{(t)} }}$ of all measurements at step $t$ is distributed~as 
\begin{gather*}
    y^{(t)} | x \sim \CN(\op^{(t)} x,\, \sigma^{2}_{y} \mI_{d_{y}}) \quad \text{with} \quad  x \sim \CN(0, \Sigma_{xx}).
\end{gather*}
Thus, $\Sigma^{(t)}_{yy} = \op^{(t)} \Sigma_{xx} (\op^{(t)})^{\top}{+}\,\sigma^{2}_{y}\mI$ is the measurement covariance and the posterior over $x$ is %
\begin{gather}
    x | y^{(t)} \sim \CN(\mu_{x|y^{(t)}}, \Sigma_{x|y^{(t)}}), \quad \text{with} \notag \\ \mu_{x|y^{(t)}}=\Sigma_{xx}(\op^{(t)})^{\top}(\Sigma_{yy}^{(t)})^{-1}y^{(t)}, \quad \text{and} \quad \Sigma_{x|y^{(t)}} = \Sigma_{xx} - \Sigma_{xx}(\op^{(t)})^{\top} (\Sigma_{yy}^{(t)})^{-1}\op^{(t)}\Sigma_{xx}.\label{eq:linear_posterior}
\end{gather}
The predictive covariance $\Sigma_{x|y^{(t)}}$ completely characterises the uncertainty of the reconstruction at step $t$ and is the building block for the angle selection criteria in \cref{subsec:linear_design}.
Note that natural images often exhibit heavy-tailed non-Gaussian statistics \citep{Seeger11scale}. Additionally, by \cref{eq:linear_posterior}, $\Sigma_{x|y^{(t)}}$ depends on the choice of angles through $\op^{(t)}$, but not on the measurements made at said angles $y^{(t)}$, precluding adaptive design. 
In \cref{subsec:prior_cov}, we construct $\Sigma_{xx}$ with correlations between nearby pixels, imitating the effects of the TV regulariser \cref{eq:TV_equation}, and with dependence on previous measurements, recovering adaptive design capability.
In the experiments, we use linear models for angle selection and afterwards we discard the predictive mean $\mu_{x|y}$ and employ the regularised approaches from \cref{sec:preliminaries} for~reconstruction.

\subsection{Experimental design with linear models} \label{subsec:linear_design}

\textbf{Acquisition objectives.} Since the linear design task is submodular \citep{Seeger2009submodular}, we greedily add one single angle per acquisition step \footnote{Submodularity guarantees this procedure obtains a score within a $(1-\nicefrac{1}{e})$ factor of the optimal strategy.}. We consider two popular acquisition objectives.

The first objective, \emph{expected information gain} (EIG) \citep{Mackay1992InformationBasedOF}, is the expected reduction in the posterior entropy $H(x|y)$ from scanning at angle $\beta$. At step $t$, it is given by
\begin{gather}\label{eq:eig}
    \text{EIG} := \mH({x} | y^{(t)}) - \BE_{p( y^{\beta} | y^{(t)})} [\mH({x} | y^{(t)}, y^{\beta})] = \logdet(\sigma^{2}_{y} \mI_{d_{\CB^{(t)}}} + \op^{\beta}\Sigma_{x | \data^{(t)}}(\op^{\beta})^{\top}) + C 
\end{gather}
where the constant $C = -\logdet(\sigma^{2}_{y} \mI)$ is independent of the angle choice. We give a derivation in \cref{appx:acq_objs} for completeness.
{Intuitively, the determinant of the matrix $\op^{\beta}\Sigma_{x | \data^{(t)}}(\op^{\beta})^{\top}\in\mathbb{R}^{d_p\times d_p}$ penalises angles for which different detector elements make correlated measurements and the log term encourages the measurements from all detector pixels to be similarly informative.}

The second objective, which we find to perform better empirically, is to choose the angles for which our prediction has the largest \emph{expected squared error} (ESE) in measurement space
\begin{gather}\label{eq:ESE}
    \text{ESE} := \BE_{p(y^{\beta}, \,x|y^{(t)})}[(y^{\beta} - \op^{\beta} x)^{\top}(y^{\beta} - \op^{\beta} x)] = \text{Tr}(\op^{\beta}\Sigma_{x | \data^{(t)}}(\op^{\beta})^{\top}) + C.
\end{gather}
This objective is equivalent to EIG in the setting where our detector has a single pixel. 

\textbf{Efficient acquisition.} 
Constructing the matrix $\op^{\beta}\Sigma_{x | \data^{(t)}}(\op^{\beta})^{\top}$ repeatedly for each candidate angle $\beta \in \bar{\CB}^{(t)}$ requires $\CO(d_{p}\cdot d_{\bar{\CB}^{(t)}})$ matrix vector products, which is very costly even for moderate size scanners.
Instead, we estimate the matrix for every angle simultaneously by drawing $K$ samples from $\CN(0, \bar{\op}^{(t)}\Sigma_{x | \data^{(t)}} (\bar{\op}^{(t)})^{\top} )$ with $\CO(K)$ matrix vector products. That is, we sample $\BR^{d_{p}\cdot d_{\bar{\CB}^{(t)}}}$ sized vectors built by concatenating the ``pseudo measurements'' for each unused angle $\beta \in \bar{\CB}^{(t)}$. We use Matheron's rule  \citep{Hoffman1991Constrained,Wilson2021Pathwise}
\begin{gather}
    \bigoplus_{\beta \in \bar{\CB}^{(t)}} \!y^{\beta}_{k} = \bar{\op}^{(t)} \!\left(x_{k} - \Sigma_{xx}(\op^{(t)})^{\top} \Sigma_{yy}^{-1}(\eta_{k} + \op^{(t)}x_{k})\right) \quad \text{with} \notag \\
     x_{k}\sim \CN(0, \Sigma_{xx}  ) \quad \text{and} \quad \eta_{k}\sim \CN(0, \sigma_{y}^{2} \mI ),\label{eq:matheron}.
\end{gather}
Here, $k \in \{1,...,K\}$ indexes different samples and $\bigoplus$ denotes vector concatenation. We compute
\begin{gather*}
    \op^{\beta}\Sigma_{x | \data^{(t)}}(\op^{\beta})^{\top} \approx K^{-1} \textstyle{\sum_{k=1}^{K}} y^{\beta}_{k} (y^{\beta}_{k})^{\top},
\end{gather*}
which is then used to estimate the acquisition objective \cref{eq:eig} or \cref{eq:ESE}. The log term makes EIG estimates only asymptotically unbiased (i.e.\ as $K\to \infty$) but we find the bias to be insignificant. Once the angle $\beta$ that maximises \cref{eq:eig} or \cref{eq:ESE} is chosen, we update $\Sigma_{yy}^{(t+1)}$ as
\begin{equation}
    \Sigma_{yy}^{(t+1)} = \begin{bmatrix}
        \Sigma_{yy}^{(t)} & \op^{(t)}\Sigma_{xx}(\op^{(t+1)})^{\top} \\
        \op^{(t+1)}\Sigma_{xx}(\op^{(t)})^{\top} & \op^{(t + 1)}\Sigma_{xx}(\op^{(t+1)})^{\top}
    \end{bmatrix},\label{eq:covariance_update}
\end{equation}
and repeat the procedure, i.e.\ return to \cref{eq:matheron}.

\begin{figure}[t]
\vspace{-1cm}
\includegraphics[width=\textwidth]{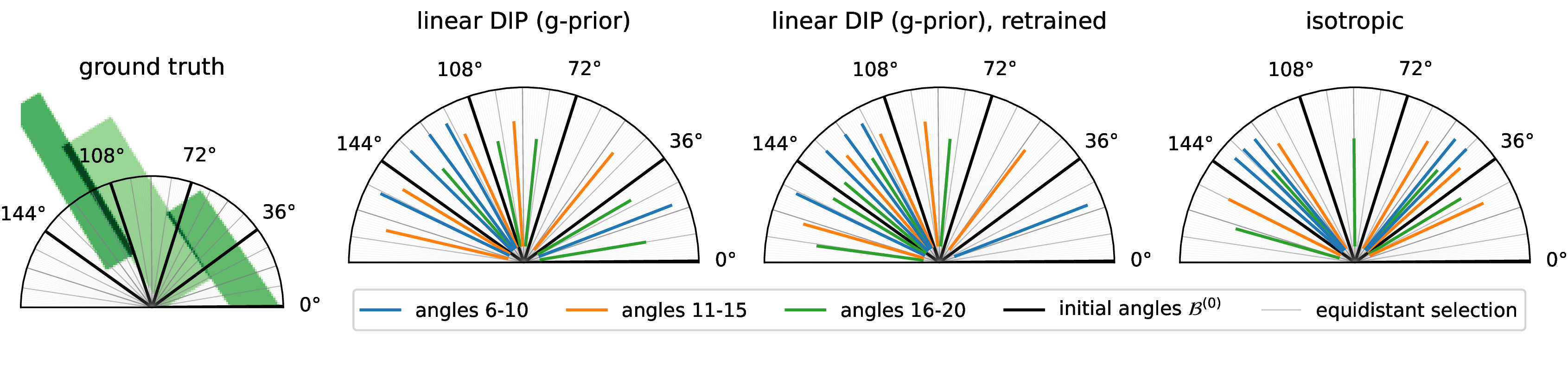}
\vspace{-1.2cm}
\caption{First 20 angles selected by each method under consideration for an example image.} \label{fig:angle_selection_image1}
\vspace{-0.3cm}
\end{figure}

\subsection{Construction of the prior covariance $\Sigma_{xx}$} \label{subsec:prior_cov}
Now we describe the construction of the Gaussian prior covariance $\Sigma_{xx}\in \BR^{d_{x} \times d_{x}}$ over reconstructions. We consider a range of models, building from very simple models to flexible data-driven ones that allows for adaptive design. 

\textbf{Isotropic model.} The simple choice $\Sigma_{xx} = \sigma^{2}_{x} I_{d_{x}}$ assumes uncorrelated pixels, and it implies a ridge regulariser for the reconstruction, which is known to perform poorly in imaging.

\textbf{Matern-$\nicefrac{1}{2}$ Process.} \cite{Antoran2022Tomography} employ the Matern-$\nicefrac{1}{2}$  covariance $[\Sigma_{xx}]_{ij,i'j'} = \sigma^{2}_{x}\exp (-\ell^{-1}\sqrt{(i-i')^{2} + (j-j')^{2}})$, where $i, j$ index the pixel locations in the image $x$, as a surrogate for TV. With the hyperparameters $\sigma^{2}_{x}$ and $\ell$ properly chosen, the prior samples and posterior inferences closely match those obtained with an intractable TV prior.

\textbf{Linearised deep image prior} \citep{barbano2022probabilistic,Antoran2022Tomography}. This data-driven prior is constructed by first fitting a DIP model on the measurements taken during the pilot scan with \cref{eq:DIP_MAP-obj}, and then adopting a linear model on the basis expansion given by the Jacobian of the trained U-net $x(\cdot)$ with respect to $\theta$ evaluated at the optimal point $\theta^{\star}$, i.e.\ $\nabla_{\theta} x(\theta)_{|\theta = \theta^{\star}} \eqqcolon \mJ \in \BR^{d_{x} \times d_{\theta}}$ \citep{Immer2021improving}. %
The resulting prior over $x$ is given by
\vspace{-0.1cm}
\begin{gather*}
    x = \mJ \theta,\quad \theta\sim \CN(0, \Sigma_{\theta}) \quad \text{and thus} \quad x \sim \CN(0, \mJ \Sigma_{\theta} \mJ^{\top}).
\end{gather*}
\vspace{-0.1cm}
The covariance $\Sigma_{xx} = \mJ \Sigma_{\theta} \mJ^{\top}$  incorporates information about the pilot measurements through the features $\mJ$. It assigns higher prior variance being near the edges in the reconstruction, cf.\ \cref{fig:variance_angles}, which are most sensitive to a change in U-net parameters. The covariance $\Sigma_{\theta} \in \BR^{d_{\theta} \times d_{\theta}}$ weights different Jacobian entries. We consider two different structures for $\Sigma_{\theta}$.
\begin{itemize}[leftmargin=*]
\vspace{-0.1cm}
  \setlength\itemsep{0pt}
     \item The filter-wise block-diagonal matrix of \cite{Antoran2022Tomography} uses a separate prior for every block in the U-net (cf.\ \cref{appx:linearised_DIP}). This choice uses a large number of hyperparameters. It risks overfitting to the pilot scan measurements resulting in uncertainty underestimation.
    \item The neural g-prior \citep{zellner_1986,antoran2022linearised} is a maximally uninformative diagonal Gaussian prior with covariance matching the diagonal of U-net's inverse Fisher information matrix, denoted $s^{-1}$, scaled by a constant $g$ (see \cref{app:gprior} for extended discussion). That~is
    \vspace{-0.3cm}
    \begin{gather*}
        \Sigma_{\theta} = g \cdot s^{-1} \mI, \quad s = d_{y^{(t)}}^{-1} \sum_{i =1}^{d_{y}} ([\op^{(t)} \mJ]_{i})^{2} \in \BR^{d_{\theta}}, \!\quad \!\text{and we choose}\! \quad \! g = (d_{y^{(0)}} d_{\theta})^{-1}\sum_{i=1}^{d_{y}}((y_{i}^{(0)})^{2} - \sigma_{y}^{2}),
    \end{gather*}
    \vspace{-0.2cm}
\end{itemize}
where $[\op \mJ]_{i}$ refers to the $i$th row of the matrix $\op \mJ$. Computing $s$ does not require measurement values and we update it every $5$ acquired angles. We compute $g$ once  using the measurements from the pilot scan.
Our choice of $s$ ensures that the Jacobian entries corresponding to all U-net weights contribute equally to the marginal prior variance over measurements. Our choice of $g$ ensures this marginal variance is equal to the empirical second moment of pilot measurements. 
\nopagebreak

All models discussed have a number of free parameters $\sigma_{y}^{2}, \sigma_{x}^{2}, \ell, \Sigma_{\theta}$, which we choose to maximise the model evidence given the pilot scan measurements. See \cref{appx:model_evidence_app} for details.

\begin{figure}[t]
\vspace{-1.0cm}
\minipage{0.329\textwidth}
  \includegraphics[width=\linewidth]{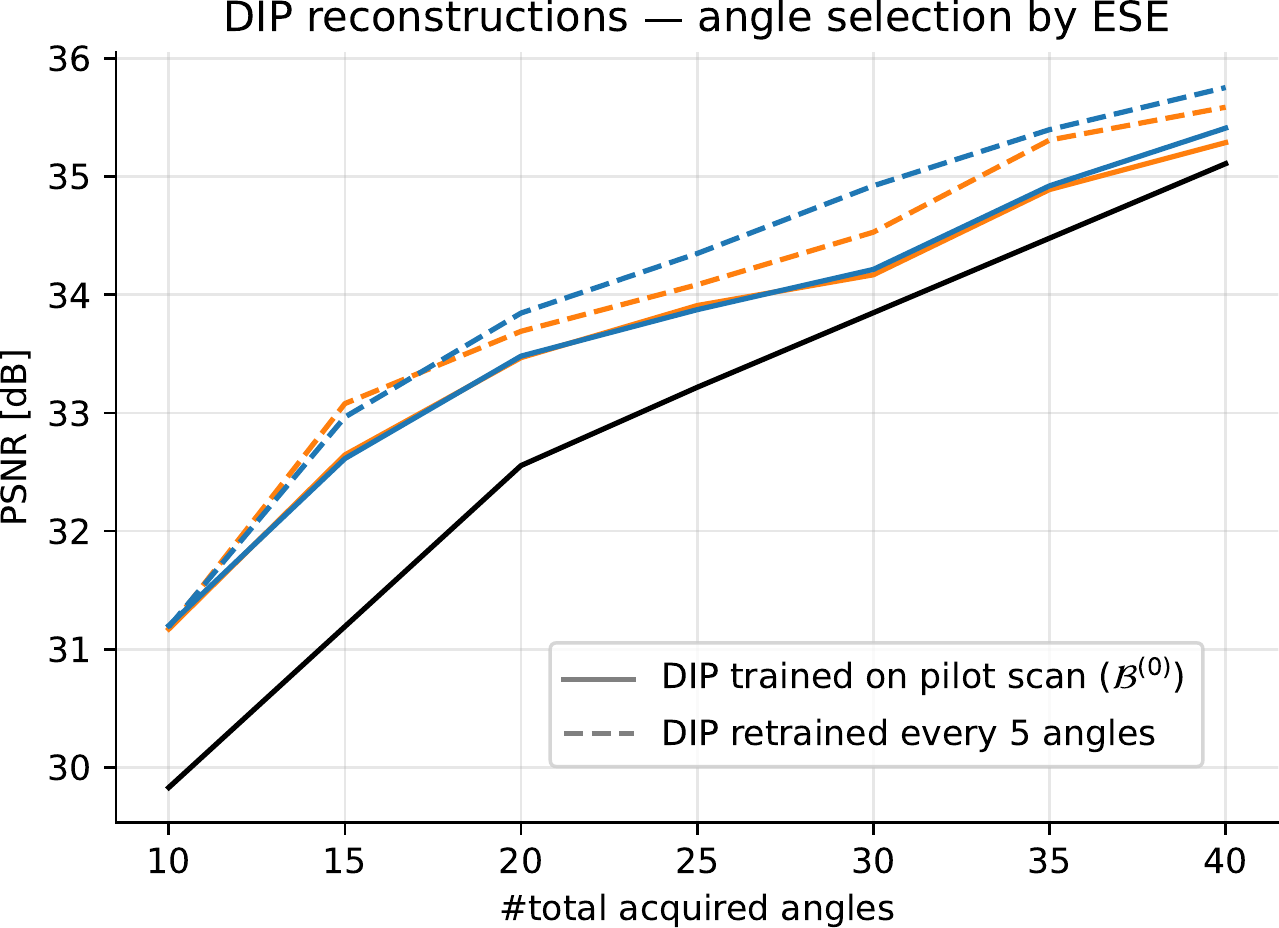}
\endminipage\hfill
\minipage{0.329\textwidth}
  \includegraphics[width=\linewidth]{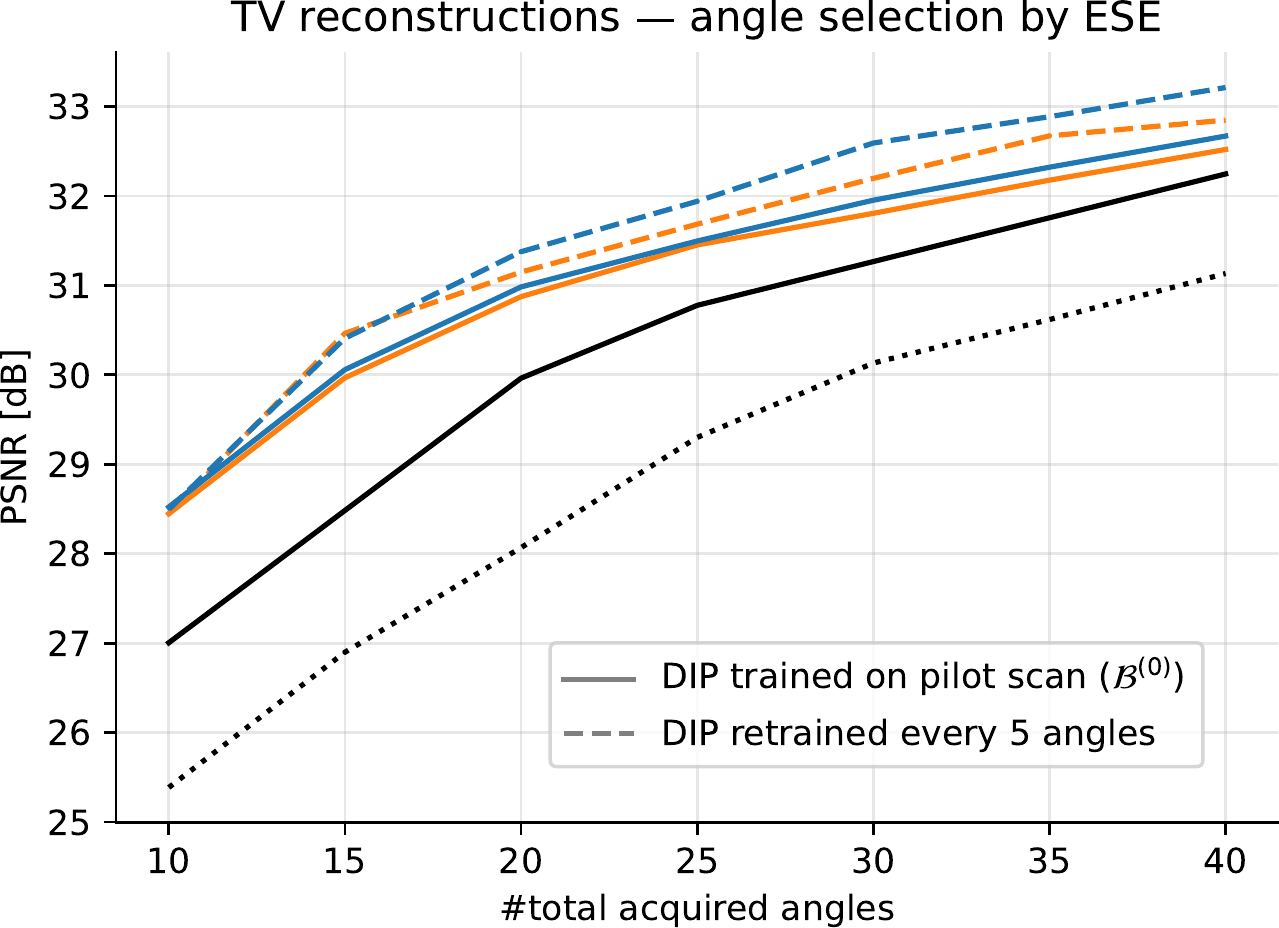}
\endminipage\hfill
\minipage{0.329\textwidth}%
  \includegraphics[width=\linewidth]{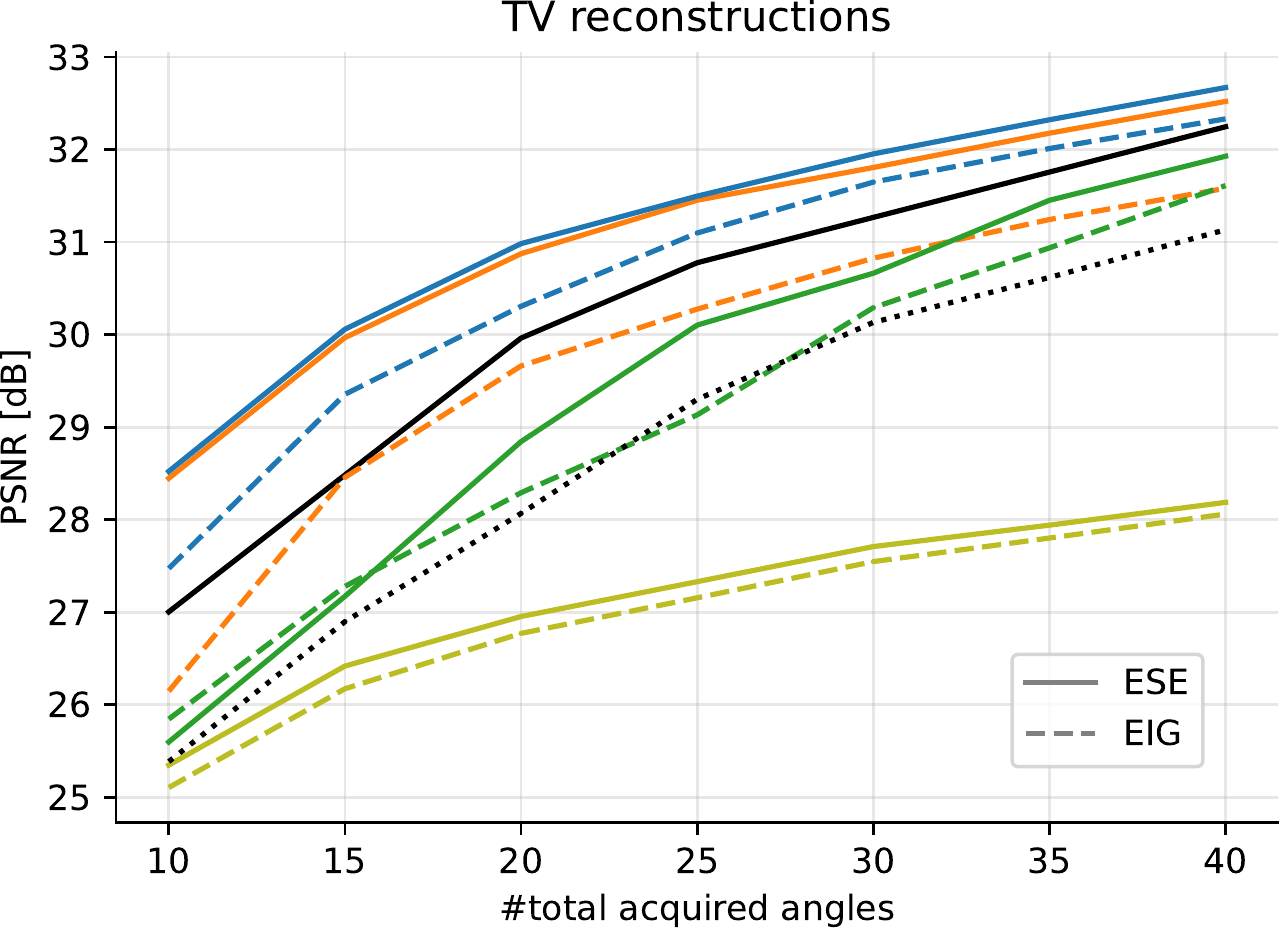}
\endminipage
\\[0.5em]
\centering
\includegraphics[width=0.85\linewidth]{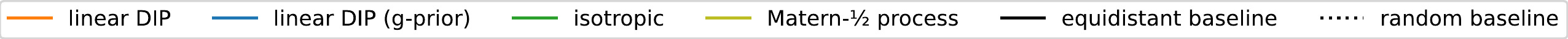}
\vspace{-0.25cm}
\caption{Reconstruction PSNR vs n. angles scanned, averaged across 30 images (5\% noise).}\label{fig:main-figure-psnr-comparison}
\vspace{-0.3cm}
\end{figure}

\section{Experiments and analysis}\label{sec:experiments}

We simulate CT measurements $y$ from $128\!\times\!128$ pixel images displaying rectangles of random proportions aligned along a randomly chosen ``preferential'' direction (see \cref{fig:angle_selection_image1} and \cref{fig:example_images}). 
The forward operator $\op$ is the discrete Radon transform, and either $5\%$ or $10\%$ white noise is added to the measurement $y$.
We divide the range $[0^\circ, 180^{\circ})$ into 200 selectable angles (i.e.\ $|\CB_{a}| = 200)$. The pilot scan measures at 5 equidistant angles, on which  we fit all models' hyperparameters and the linearised DIP's U-net (see \cref{appx:model_evidence_app}).
Then, we apply the methods in \cref{subsec:linear_design} to produce designs consisting of 35 additional angles.
For every 5 acquired angles, we evaluate reconstruction quality using both the DIP (i.e.\ \cref{eq:DIP_MAP-obj}), and the traditional TV regularised approach (i.e.\ \cref{eq:simple_optimisation_objective}). We include equidistant and random angle selection as baselines. On an NVIDIA A100 GPU, a full linearised DIP acquisition step with $K=3000$ samples takes 9 seconds and the full design takes 5 minutes.
\Cref{appx:full_setup} contains full experimental details.

For the \textbf{linearised DIP}, we consider training our U-net and prior hyperparameters only on the pilot scan, and also retraining every 5 angles. \Cref{fig:angle_selection_image1} shows both approaches can identify and prioritise the preferential direction, leading to reconstructions that \emph{outperform the equidistant angle baseline by over 1.5 dB} in the range of $[10, 15]$ angles (see \cref{fig:main-figure-psnr-comparison}). 
During this initial stage, the linearised DIP requires roughly \emph{30\% less scanned angles} to match the equidistant baseline's performance.
The performance gap decreases as we select more angles, although linearised DIP remains more efficient even after 40 angles. 
Retraining the U-net provides most benefits in the large angle regime. It increases focus on preferential directions and consistently provides gains ${>}0.5$dB after 20 angles. All gains over the equidistant baseline are obtained with both DIP (i.e.\ \cref{eq:DIP_MAP-obj}) and traditional TV regularised reconstruction (i.e.\ \cref{eq:simple_optimisation_objective}). In the high 10\% noise setting, gains from experimental design are smaller, but still significant (see~\cref{appx:add_exp_results}). 

The \textbf{isotropic and Matern-$\nicefrac{1}{2}$} models' uncertainty estimates are independent of the pilot measurements. These models prioritise clustered sets of oblique angles which maximise the length of quanta trajectories in the image. They perform similar to or worse than random. We explore this negative result in \cref{appx:add_exp_results}, finding it due to overfitting of hyperparameters.

\textbf{ESE outperforms EIG} across models. For the linearised DIP, this gap is smaller when using the g-prior. 
We hypothesise that model misspecification and hyperparameter overfitting may result in poor measurement covariance estimates, in turn degrading EIG estimates.

\section{Conclusion and future work}\label{test}

Our results suggest that dependence on the measurement data, i.e.\ adaptivity, is key to outperforming equidistant angle selection in CT reconstruction, a notoriously difficult task \citep{Shen2022learningtoscan,Helin2021GaussianTV}. Distinctly from previous work, our methods only necessitate a pilot scan instead of being fully online, increasing applicability. We observe the largest gains in the 10 to 20 angle regime, where our designs reduce the angle requirement by roughly 30\% without loss of reconstruction quality. This is true for both traditional TV-regularised and DIP reconstructions.
In future, we aim to apply linearised DIP designs to real measurements.

\acks{ We would like to thank Eric Nalisnick and Mark van der Wilk for helpful discussions. R.B. acknowledges support from the i4health PhD studentship
(UK EPSRC EP/S021930/1), and from The Alan Turing Institute (UK EPSRC EP/N510129/1). The work of B.J. is partially supported by UK EPSRC grants EP/T000864/1 and EP/V026259/1.
J.L. is funded by the German Research Foundation (DFG; GRK 2224/1), and additionally acknowledges support from the DELETO project funded by the Federal Ministry of Education and Research (BMBF, project
number 05M20LBB). J.A. acknowledges support from Microsoft Research, through its PhD Scholarship Programme, and from the EPSRC. J.A. also acknowledges travel support from ELISE (GA no 951847). This work has been performed using resources provided by the Cambridge Tier-2 system operated by the University of Cambridge Research Computing Service (http://www.hpc.cam.ac.uk) funded by EPSRC Tier-2 capital grant EP/T022159/1.
}

\vskip 0.2in
\bibliography{sample}

\begin{thebibliography}{34}
\providecommand{\natexlab}[1]{#1}
\providecommand{\url}[1]{\texttt{#1}}
\expandafter\ifx\csname urlstyle\endcsname\relax
  \providecommand{\doi}[1]{doi: #1}\else
  \providecommand{\doi}{doi: \begingroup \urlstyle{rm}\Url}\fi

\bibitem[{Antoran} and {Miguel}(2019)]{antoran2019disentangling}
J.~{Antoran} and A.~{Miguel}.
\newblock Disentangling and learning robust representations with natural
  clustering.
\newblock In \emph{2019 18th IEEE International Conference On Machine Learning
  And Applications (ICMLA)}, pages 694--699, 2019.

\bibitem[Antor{\'{a}}n et~al.(2020)Antor{\'{a}}n, Allingham, and
  Hern{\'{a}}ndez{-}Lobato]{Antoran20depth}
Javier Antor{\'{a}}n, James~Urquhart Allingham, and Jos{\'{e}}~Miguel
  Hern{\'{a}}ndez{-}Lobato.
\newblock Depth uncertainty in neural networks.
\newblock In Hugo Larochelle, Marc'Aurelio Ranzato, Raia Hadsell,
  Maria{-}Florina Balcan, and Hsuan{-}Tien Lin, editors, \emph{Advances in
  Neural Information Processing Systems 33: Annual Conference on Neural
  Information Processing Systems 2020, NeurIPS 2020, December 6-12, 2020,
  virtual}, 2020.
\newblock URL
  \url{https://proceedings.neurips.cc/paper/2020/hash/781877bda0783aac5f1cf765c128b437-Abstract.html}.

\bibitem[Antor{\'{a}}n et~al.(2021)Antor{\'{a}}n, Bhatt, Adel, Weller, and
  Hern{\'{a}}ndez{-}Lobato]{Antoran21clue}
Javier Antor{\'{a}}n, Umang Bhatt, Tameem Adel, Adrian Weller, and
  Jos{\'{e}}~Miguel Hern{\'{a}}ndez{-}Lobato.
\newblock Getting a {CLUE:} {A} method for explaining uncertainty estimates.
\newblock In \emph{9th International Conference on Learning Representations,
  {ICLR} 2021, Virtual Event, Austria, May 3-7, 2021}. OpenReview.net, 2021.
\newblock URL \url{https://openreview.net/forum?id=XSLF1XFq5h}.

\bibitem[Antoran et~al.(2022)Antoran, Allingham, Janz, Daxberger, Nalisnick,
  and Hern{\'a}ndez-Lobato]{antoran2022linearised}
Javier Antoran, James~Urquhart Allingham, David Janz, Erik Daxberger, Eric
  Nalisnick, and Jos{\'e}~Miguel Hern{\'a}ndez-Lobato.
\newblock Linearised laplace inference in networks with normalisation layers
  and the neural g-prior.
\newblock In \emph{Fourth Symposium on Advances in Approximate Bayesian
  Inference}, 2022.
\newblock URL \url{https://openreview.net/forum?id=uUH8x-h9zdB}.

\bibitem[Antor\'{a}n et~al.(2022)Antor\'{a}n, Barbano, Leuschner,
  Hern\'{a}ndez-Lobato, and Jin]{Antoran2022Tomography}
Javier Antor\'{a}n, Riccardo Barbano, Johannes Leuschner, Jos\'{e}~Miguel
  Hern\'{a}ndez-Lobato, and Bangti Jin.
\newblock A probabilistic deep image prior for computational tomography.
\newblock Preprint, arXiv:2203.00479, 2022.

\bibitem[Antor{\'{a}}n et~al.(2022)Antor{\'{a}}n, Janz, Allingham, Daxberger,
  Barbano, Nalisnick, and Hern{\'{a}}ndez{-}Lobato]{antoran2022adapting}
Javier Antor{\'{a}}n, David Janz, James~Urquhart Allingham, Erik~A. Daxberger,
  Riccardo Barbano, Eric~T. Nalisnick, and Jos{\'{e}}~Miguel
  Hern{\'{a}}ndez{-}Lobato.
\newblock Adapting the linearised laplace model evidence for modern deep
  learning.
\newblock \emph{CoRR}, abs/2206.08900, 2022.
\newblock \doi{10.48550/arXiv.2206.08900}.
\newblock URL \url{https://doi.org/10.48550/arXiv.2206.08900}.

\bibitem[Baguer et~al.(2020)Baguer, Leuschner, and Schmidt]{baguer2020diptv}
Daniel~Otero Baguer, Johannes Leuschner, and Maximilian Schmidt.
\newblock Computed tomography reconstruction using deep image prior and learned
  reconstruction methods.
\newblock \emph{Inverse Problems}, 36\penalty0 (9):\penalty0 094004, 2020.

\bibitem[Barbano et~al.(2021)Barbano, Leuschner, Schmidt, Denker, Hauptmann,
  Maa{\ss}, and Jin]{barbano2021deep}
Riccardo Barbano, Johannes Leuschner, Maximilian Schmidt, Alexander Denker,
  Andreas Hauptmann, Peter Maa{\ss}, and Bangti Jin.
\newblock Is deep image prior in need of a good education?
\newblock \emph{arXiv preprint arXiv:2111.11926}, 2021.

\bibitem[Barbano et~al.(2022)Barbano, Antor{\'e}n, Hern{\'a}ndez-Lobato, and
  Jin]{barbano2022probabilistic}
Riccardo Barbano, Javier Antor{\'e}n, Jos{\'e}~Miguel Hern{\'a}ndez-Lobato, and
  Bangti Jin.
\newblock A probabilistic deep image prior over image space.
\newblock In \emph{Fourth Symposium on Advances in Approximate Bayesian
  Inference}, 2022.
\newblock URL \url{https://openreview.net/forum?id=qtFPfwJWowM}.

\bibitem[Burger et~al.(2021)Burger, Hauptmann, Helin, Hyv\"{o}nen, and
  Puska]{Burger2021lineardesign}
Martin Burger, Andreas Hauptmann, Tapio Helin, Nutti Hyv\"{o}nen, and
  Juha-Pekka Puska.
\newblock Sequentially optimized projections in x-ray imaging.
\newblock \emph{Inverse Problems}, 37\penalty0 (7):\penalty0 075006, 2021.
\newblock \doi{10.1088/1361-6420/ac01a4}.

\bibitem[Buzug(2011)]{buzug2011computed}
Thorsten~M Buzug.
\newblock Computed tomography.
\newblock In \emph{Springer handbook of medical technology}, pages 311--342.
  Springer, 2011.

\bibitem[Chaloner and Verdinelli(1995)]{Chaloner1995review}
Kathryn Chaloner and Isabella Verdinelli.
\newblock {Bayesian experimental design: A review}.
\newblock \emph{Statistical Science}, 10\penalty0 (3):\penalty0 273 -- 304,
  1995.
\newblock \doi{10.1214/ss/1177009939}.
\newblock URL \url{https://doi.org/10.1214/ss/1177009939}.

\bibitem[Chambolle et~al.(2010)Chambolle, Caselles, Cremers, Novaga, and
  Pock]{chambolle2010introduction}
Antonin Chambolle, Vicent Caselles, Daniel Cremers, Matteo Novaga, and Thomas
  Pock.
\newblock An introduction to total variation for image analysis.
\newblock In \emph{Theoretical foundations and numerical methods for sparse
  recovery}, pages 263--340. de Gruyter, 2010.

\bibitem[Daxberger et~al.(2021)Daxberger, Nalisnick, Allingham, Antor{\'{a}}n,
  and Hern{\'{a}}ndez{-}Lobato]{Daxberger21subnetwork}
Erik~A. Daxberger, Eric~T. Nalisnick, James~Urquhart Allingham, Javier
  Antor{\'{a}}n, and Jos{\'{e}}~Miguel Hern{\'{a}}ndez{-}Lobato.
\newblock Bayesian deep learning via subnetwork inference.
\newblock In Marina Meila and Tong Zhang, editors, \emph{Proceedings of the
  38th International Conference on Machine Learning, {ICML} 2021, 18-24 July
  2021, Virtual Event}, volume 139 of \emph{Proceedings of Machine Learning
  Research}, pages 2510--2521. {PMLR}, 2021.
\newblock URL \url{http://proceedings.mlr.press/v139/daxberger21a.html}.

\bibitem[Fedorov(1972)]{Fedorov1972theory}
V.~V. Fedorov.
\newblock \emph{{Theory of Optimal Experiments}}.
\newblock Academic Press New York, 1972.
\newblock ISBN 0122507509.

\bibitem[Feng(2015)]{chi2015modelerror}
Chi Feng.
\newblock \emph{Optimal Bayesian experimental design in the presence of model
  error}.
\newblock PhD thesis, Massachusetts Institute of Technology, Cambridge, MA,
  2015.

\bibitem[Foster(2021)]{foster2021Design}
Adam~Evan Foster.
\newblock \emph{Variational, Monte Carlo and Policy-Based Approaches to
  Bayesian Experimental Design}.
\newblock PhD thesis, University of Oxford, 2021.

\bibitem[Helin et~al.(2022)Helin, Hyv\"{o}nen, and Puska]{Helin2021GaussianTV}
Tapio Helin, Nuutti Hyv\"{o}nen, and Juha-Pekka Puska.
\newblock Edge-promoting adaptive {B}ayesian experimental design for {X}-ray
  imaging.
\newblock \emph{SIAM J. Sci. Comput.}, 44\penalty0 (3):\penalty0 B506--B530,
  2022.
\newblock ISSN 1064-8275.
\newblock \doi{10.1137/21M1409330}.

\bibitem[{Hoffman} and {Ribak}(1991)]{Hoffman1991Constrained}
Yehuda {Hoffman} and Erez {Ribak}.
\newblock Constrained realizations of {G}aussian fields: a simple algorithm.
\newblock \emph{Astrophys. J. Lett.}, 380:\penalty0 L5--L8, 1991.
\newblock \doi{10.1086/186160}.

\bibitem[Immer et~al.(2021{\natexlab{a}})Immer, Bauer, Fortuin, R{\"{a}}tsch,
  and Khan]{Immer21Selection}
Alexander Immer, Matthias Bauer, Vincent Fortuin, Gunnar R{\"{a}}tsch, and
  Mohammad~Emtiyaz Khan.
\newblock Scalable marginal likelihood estimation for model selection in deep
  learning.
\newblock In Marina Meila and Tong Zhang, editors, \emph{Proceedings of the
  38th International Conference on Machine Learning, {ICML} 2021, 18-24 July
  2021, Virtual Event}, volume 139 of \emph{Proceedings of Machine Learning
  Research}, pages 4563--4573. {PMLR}, 2021{\natexlab{a}}.
\newblock URL \url{http://proceedings.mlr.press/v139/immer21a.html}.

\bibitem[Immer et~al.(2021{\natexlab{b}})Immer, Korzepa, and
  Bauer]{Immer2021improving}
Alexander Immer, Maciej Korzepa, and Matthias Bauer.
\newblock Improving predictions of bayesian neural nets via local
  linearization.
\newblock In Arindam Banerjee and Kenji Fukumizu, editors, \emph{The 24th
  International Conference on Artificial Intelligence and Statistics, {AISTATS}
  2021, April 13-15, 2021, Virtual Event}, volume 130 of \emph{Proceedings of
  Machine Learning Research}, pages 703--711. {PMLR}, 2021{\natexlab{b}}.
\newblock URL \url{http://proceedings.mlr.press/v130/immer21a.html}.

\bibitem[Ito and Jin(2014)]{ito2014inverse}
Kazufumi Ito and Bangti Jin.
\newblock \emph{Inverse problems: Tikhonov theory and algorithms}, volume~22.
\newblock World Scientific, 2014.

\bibitem[Lee et~al.(2020)Lee, Schoenholz, Pennington, Adlam, Xiao, Novak, and
  Sohl-Dickstein]{lee2020finite}
Jaehoon Lee, Samuel Schoenholz, Jeffrey Pennington, Ben Adlam, Lechao Xiao,
  Roman Novak, and Jascha Sohl-Dickstein.
\newblock Finite versus infinite neural networks: an empirical study.
\newblock \emph{Advances in Neural Information Processing Systems},
  33:\penalty0 15156--15172, 2020.

\bibitem[Mackay(1992{\natexlab{a}})]{Mackay1992InformationBasedOF}
David J.~C. Mackay.
\newblock Information-based objective functions for active data selection.
\newblock \emph{Neural Computation}, 4:\penalty0 590--604, 1992{\natexlab{a}}.

\bibitem[Mackay(1992{\natexlab{b}})]{Mackay1992Thesis}
David John~Cameron Mackay.
\newblock \emph{Bayesian Methods for Adaptive Models}.
\newblock PhD thesis, USA, 1992{\natexlab{b}}.
\newblock UMI Order No. GAX92-32200.

\bibitem[Ronneberger et~al.(2015)Ronneberger, Fischer, and
  Brox]{ronneberger2015u}
Olaf Ronneberger, Philipp Fischer, and Thomas Brox.
\newblock U-net: Convolutional networks for biomedical image segmentation.
\newblock In \emph{International Conference on Medical Image Computing and
  Computer-Assisted Intervention}, pages 234--241. Springer, 2015.

\bibitem[Rudin et~al.(1992)Rudin, Osher, and Fatemi]{rudin1992nonlinear}
Leonid~I Rudin, Stanley Osher, and Emad Fatemi.
\newblock Nonlinear total variation based noise removal algorithms.
\newblock \emph{Physica D: nonlinear phenomena}, 60\penalty0 (1-4):\penalty0
  259--268, 1992.

\bibitem[Seeger(2009)]{Seeger2009submodular}
Matthias~W. Seeger.
\newblock On the submodularity of linear experimental design.
\newblock Technical report, 2009.
\newblock URL \url{http://infoscience.epfl.ch/record/175483}.

\bibitem[Seeger and Nickisch(2011)]{Seeger11scale}
Matthias~W. Seeger and Hannes Nickisch.
\newblock Large scale bayesian inference and experimental design for sparse
  linear models.
\newblock \emph{{SIAM} J. Imaging Sci.}, 4\penalty0 (1):\penalty0 166--199,
  2011.
\newblock \doi{10.1137/090758775}.

\bibitem[Shen et~al.(2022)Shen, Wang, Wu, Yang, and
  Dong]{Shen2022learningtoscan}
Ziju Shen, Yufei Wang, Dufan Wu, Xu~Yang, and Bin Dong.
\newblock Learning to scan: A deep reinforcement learning approach for
  personalized scanning in ct imaging.
\newblock \emph{Inverse Problems and Imaging}, 16\penalty0 (1):\penalty0
  179--195, 2022.

\bibitem[Tikhonov and Arsenin(1977)]{tikhonov1977solutions}
Andrey~N. Tikhonov and Vasiliy~Y. Arsenin.
\newblock \emph{Solutions of ill-posed problems}.
\newblock V. H. Winston \& Sons, Washington, D.C.: John Wiley \& Sons, New
  York, 1977.
\newblock Translated from the Russian, Preface by translation editor Fritz
  John, Scripta Series in Mathematics.

\bibitem[Ulyanov et~al.(2020)Ulyanov, Vedaldi, and Lempitsky]{ulyanov2020dip}
Dmitry Ulyanov, Andrea Vedaldi, and Victor Lempitsky.
\newblock Deep image prior.
\newblock \emph{Int. J. Comput. Vis.}, 128\penalty0 (7):\penalty0 1867--1888,
  2020.
\newblock ISSN 1573-1405.
\newblock \doi{10.1007/s11263-020-01303-4}.

\bibitem[Wilson et~al.(2021)Wilson, Borovitskiy, Terenin, Mostowsky, and
  Deisenroth]{Wilson2021Pathwise}
James~T. Wilson, Viacheslav Borovitskiy, Alexander Terenin, Peter Mostowsky,
  and Marc~Peter Deisenroth.
\newblock Pathwise conditioning of gaussian processes.
\newblock \emph{J. Mach. Learn. Res.}, 22:\penalty0 105:1--105:47, 2021.
\newblock URL \url{http://jmlr.org/papers/v22/20-1260.html}.

\bibitem[Zellner(1986)]{zellner_1986}
Arnold Zellner.
\newblock \emph{On assessing prior distributions and Bayesian regression
  analysis with g prior distributions}, volume~6 of \emph{Studies in Bayesian
  Econometrics and Statistics}, pages 233--243.
\newblock Elsevier, 1986.

\end{thebibliography}
\clearpage

\renewcommand{\theHsection}{A\arabic{section}}

\appendix

\section{Discussion on acquisition objectives}\label{appx:acq_objs}

The descriptions that follow are well known within the experimental design community but may be of interest to readers with a background in CT. Thus we describe these facts for the convenience of readers, and refer readers to \cite{Fedorov1972theory,Chaloner1995review} for a comprehensive introduction to experimental design and to \cite{Mackay1992InformationBasedOF} for a Bayesian perspective on experimental design. 

EIG quantifies the information (in nats) we expect to gain by observing the detector elements' measurements for an angle or set of angles \citep{Mackay1992InformationBasedOF}. Since our experiments employ greedy angle selection, we derive EIG for measurements at a single angle $\beta$. The generalisation to the multi-angle setting is straightforward. EIG is the expected decrease in posterior entropy from observing the detector elements' measurements at $\beta$:
\begin{gather*}
    \text{EIG} = \mH({x} | \data^{(t-1)}) - \BE_{p({\data}^{\beta} | \data^{(t-1)})} [\mH({x} | \data^{(t-1)}, y^{\beta})],
\end{gather*}
where we take an expectation over the new measurement $y^{\beta}$, since EIG is computed before the measurement $y^\beta$ is made. For this, we use the posterior predictive distribution $p({\data}^{\beta} | \data^{(t-1)})$ given our previous measurements $\data^{(t-1)}$
\begin{gather*}
    p({\data}^{\beta} | \data^{(t-1)}) = \int p({\data}^{\beta} | x) p(x|y^{(t-1)})\, {\rm d}x.
\end{gather*}
For the linear-Gaussian case, this integral can be evaluated in closed form, although this will not be necessary for our purposes.

EIG is also equal to the mutual information $MI({x}, {\data}^{\beta} | \data^{(t-1)})$ between the reconstruction $x$ and the new measurement $y^\beta$ conditional on the previous measurements $\data^{(t-1)}$, giving an interpretation as aiming to select the angle $\beta$ most informative towards the reconstruction. 
For fixed model hyperparameters, EIG is always greater or equal than $0$ since making additional measurements cannot increase the uncertainty in the reconstruction. 

The entropy of a multivariate Gaussian $\mathcal{N}(\mu, \Sigma)$ is $\mH = \frac{1}{2} \logdet(\Sigma) + \frac{d}{2}(\log(2 \pi) + 1)$. For a fixed dimensionality $d$, the second term is constant across design steps and thus we only need to focus on the log determinant. The entropy does not depend on the distribution mean but only its covariance.
Thus, taking $y^{(t)} = [y^{(t-1)}, y^{\beta}]$, we can write
\begin{gather*}
\text{EIG} = \logdet(\Sigma_{x | \data^{(t-1)}}) -  \logdet(\Sigma_{x | \data^{(t)}}).
\end{gather*}
Since the covariance $\Sigma_{x | \data^{(t-1)}}$ does not depend on the new angle choice $\beta$, maximising EIG is equivalent to choosing the angle which minimises the updated covariance log-determinant $\logdet(\Sigma_{x | \data^{(t)}})$. Hence, the EIG objective for linear models is also known as the D(eterminant)-optimal criterion.

We can obtain a more convenient expression for EIG by noting the sequential nature of Bayesian learning; when data is observed, the prior is updated to a posterior. This posterior represents the updated beliefs and, as such, acts as a prior distribution for further inferences
\begin{gather*}
p(x | y^{(t)}) = \frac{p(y^{\beta} | x) p(x | y^{(t-1)}) }{p(y^{\beta} | y^{(t-1)})}.
\end{gather*}
For conjugate Gaussian-linear models, we can apply this principle to obtain the posterior covariance at time $t$ from the covariance at time $t-1$ using the matrix determinant lemma
\begin{gather*}
\logdet(\Sigma_{x | \data^{(t)}}) = -  \logdet(\Sigma_{x | \data^{(t-1)}}^{-1}) - \logdet(\sigma^{-2}_{y} \mI) - \logdet(\sigma^{2}_{y} \mI + \op^{(t)}_{0}\Sigma_{x | \data^{(t-1)}}\op^{\top,(t)}_{0}).
\end{gather*}
Thus, we have
\begin{align*}
    \text{EIG} &= \logdet(\Sigma_{x | \data^{(t-1)}}) -  \logdet(\Sigma_{x | \data^{(t)}})  \\ &=\logdet(\Sigma_{x | \data^{(t-1)}}) - [-  \logdet(\Sigma_{x | \data^{(t-1)}}^{-1}) - \logdet(\sigma^{-2}_{y} \mI) - \logdet(\sigma^{2}_{y} \mI + \op^{\beta}\Sigma_{x | \data^{(t-1)}}(\op^{\beta})^{\top})] \\
    &=-\logdet(\sigma^{2}_{y} \mI) + \logdet(\sigma^{2}_{y} \mI + \op^{\beta}\Sigma_{x | \data^{(t-1)}}(\op^{\beta})^{\top})\\
    &= \logdet(\sigma^{2}_{y} \mI + \op^{\beta}\Sigma_{x | \data^{(t-1)}}(\op^{\beta})^{\top}) + C
\end{align*}
where the constant $C = -\logdet(\sigma^{2}_{y} \mI)$ is independent of angle choice, yielding the objective we use for angle selection in practise.

The ESE objective in \cref{eq:ESE} aims to minimise the squared prediction error in measurement space. Objectives of this kind are commonly known as (A)verage-optimal. However, ESE is A-optimal over measurement space $y$, not over image space $x$. ESE is crucially different from minimising the arguably more relevant expected squared reconstruction error, a more computationally expensive criterion. ESE can be understood as a naive simplification of EIG, by discarding correlations between detector pixels, making $\logdet(\op^{\beta}\Sigma_{x | \data^{(t-1)}}(\op^{\beta})^{\top})$ match $\sum_{i<d_{p}} \log [\op^{\beta}\Sigma_{x | \data^{(t-1)}}(\op^{\beta})^{\top}]_{ii}$. Then, the order of log and sum are switched, something that will only be true if every element under the sum is the same.
Having reached this point, since the log function is monotonic, it does not affect angle selection and the criterion matches the trace of $\op^{\beta}\Sigma_{x | \data^{(t-1)}}(\op^{\beta})^{\top}$.

\nocite{Antoran20depth}
\nocite{Antoran21clue}
\nocite{Daxberger21subnetwork}
\nocite{antoran2019disentangling}

\section{Hyperparameter selection via model evidence maximisation}\label{appx:model_evidence_app}

For the conjugate linear-Gaussian model, the model evidence can be computed in closed form
\begin{gather*}
    \log p(y) = \log \CN(y; 0, \Sigma_{yy}) = - \frac{1}{2} \left( y^{\top} \Sigma_{yy}^{-1} y + \logdet(\Sigma_{yy}) \right) + C \\
    \text{with} \quad \Sigma_{yy} = \op \Sigma_{xx} \op^{\top} + \sigma_{y}^{2} I_{d_{y}}
\end{gather*}
and $C=\nicefrac{-d_{y}}{2}  \log 2 \pi$. This expression is straightforward to compute for the isotropic and Matern-$\nicefrac{1}{2}$ models.
The linear solve against $\Sigma_{yy}$ and log-determiant operations, while costly, are tractable to perform when the dimensionality of $y$ is low. This is the case in our experimental setup, where we use the measurements from our 5 angle pilot scan $y^{(0)}$, specifically, $d_{y}= d_{p}\cdot d_{\CB} = 5 \cdot 183 = 915$. 
We refer to \cite{Antoran2022Tomography} for discussion of efficient computation of the model evidence for the linearised DIP. For additional discussion on the motivation for the model evidence objective, its applications and pitfalls, we refer to \cite{Mackay1992Thesis,Immer21Selection,antoran2022adapting}.

Selecting prior hyperparameters with the model evidence is often claimed to be immune from overfitting due to the flexibility of the prior model being relatively low. However, when the number of measurements is small, e.g.\ after performing the pilot scan, overfitting is still possible. Indeed we observe the Matern-$\nicefrac{1}{2}$ model suffers due to this issue in our experiments. We further discuss this in \cref{appx:add_exp_results}.

The risk of overfitting is also high for the linearised DIP model of \cite{barbano2022probabilistic,Antoran2022Tomography}. Here, the basis expansion is selected by training a U-net on the pilot measurements and the number of hyperparameters is twice the number of U-net blocks, making this prior class very flexible. 
This has motivated the use of the neural g-prior \citep{antoran2022linearised}, discussed in the following section.

\section{Discussion on the neural g-prior}\label{app:gprior}

The neural g-prior $\Sigma_{\theta} = g \cdot s^{-1} \mI$ was introduced by \citet{antoran2022linearised} as an approach to ``normalise'' the second moment of the Jacobian feature expansion analogously to standard data normalisation. This normalisation ensures that the Jacobian entries corresponding to all network weights contribute equally to the predictions at the train points, or in our case, to the predictions at the already measured angles. We refer to \citet{antoran2022linearised} for a full derivation.

\citet{antoran2022linearised} learn the variance scale $g$ with the model evidence objective. However, it is well known that this procedure can overfit in the small-data regime. To prevent overfitting, in this work we choose $g$ using the heuristic
\begin{gather*}
 g = (d_{y} d_{\theta})^{-1}\sum_{i=1}^{d_{y}}((y_{i})^{2} - \sigma_{y}^{2}).
\end{gather*}
This choice is made so that the marginal predictive variance averaged across measurement locations matches the empirical second moment of the observed targets, which we will denote $\BE[y^2] = d_{y}^{-1}\sum_{i=1}^{d_{y}}y_{i}^{2}$. In other words, when using this prior over weights, our prior over measurements will have roughly the ``right'' variance.
To see this, first recall
\begin{gather*}
s = d_{y}^{-1} \sum_{i =1}^{d_{y}} ([\op \mJ]_{i})^{2} 
\end{gather*}
where $[\op \mJ]_{i}$ refers to the $i$th row of the matrix $\op \mJ$ and we will use $[\op \mJ]_{ij}$ to index each scalar entry of this matrix. We now expand the average marginal variance across measurements when using the neural g-prior
\begin{align*}
d_{y}^{-1} \sum_{i=1}^{d_y} [\Sigma_{yy}]_{ii} &= d_{y}^{-1} \sum_{i=1}^{d_y} [\op \mJ (g s^{-1} I ) \op^{\top} \mJ^{\top}]_{ii} + \sigma_{y}^{2}\\
&= g d_{y}^{-1} \sum_{i=1}^{d_y} s^{-1}_{i} \sum^{d_{\theta}}_{j} [\op \mJ]^{2}_{ij} + \sigma_{y}^{2}\\
&= (\BE[y^2] - \sigma_{y}^{2}) d_{\theta}^{-1} \sum_{i=1}^{d_y} \sum^{d_{\theta}}_{j=1}  \frac{ [\op \mJ]_{ij}^{2} }{ \sum^{d_{y}}_{k=1} [\op \mJ]_{kj}^{2}} + \sigma_{y}^{2} \\
&= (\BE[y^2] - \sigma_{y}^{2}) d_{\theta}^{-1} \sum^{d_{\theta}}_{j=1}  \frac{ \sum_{i=1}^{d_y} [\op \mJ]_{ij}^{2} }{ \sum^{d_{y}}_{k=1} [\op \mJ]_{kj}^{2}} + \sigma_{y}^{2}\\
&=  (\BE[y^2] - \sigma_{y}^{2}) \left( d_{\theta}^{-1} \sum^{d_{\theta}}_{j=1} 1 \right)+ \sigma_{y}^2 = \BE[y^2],
\end{align*}
showing the property.

For \cite{Antoran2022Tomography}, model evidence optimisation is the most computationally costly step of inference. Avoiding model evidence optimisation speeds up inference and thus angle selection, making our proposed procedure more attractive for a real deployment.

Additionally, in \cref{fig:main-figure-psnr-comparison}, we observe that the EIG objective performs best when combined with the neural g-prior. Arguably, EIG is a better motivated selection criterion than ESE but performs worse than equidistant selection when combined with all models except the g-prior linearised DIP. We hypothesise that model misspecification introduces error in our estimates of relative marginal variances and covariances across detector pixel measurements, in turn degrading the performance of EIG. ESE is less sensitive to these, as discussed in \cref{appx:acq_objs}. Since the neural g-prior is a maximally uninformative prior, it somewhat mitigates model misspecification, improving the performance of EIG acquisition.

\section{Full experimental setup}\label{appx:full_setup}

\subsection{Dataset generation} 
\begin{figure}[t]
    \centering
    \includegraphics[width=0.22\linewidth]{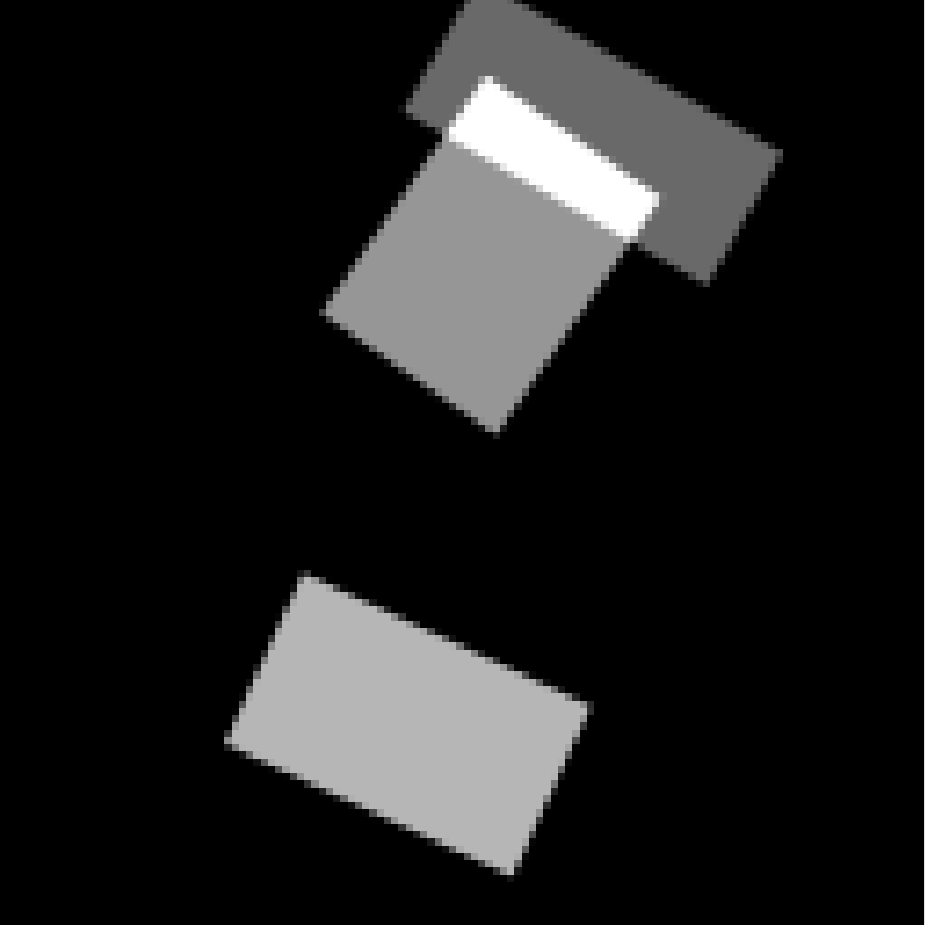}
    \includegraphics[width=0.22\linewidth]{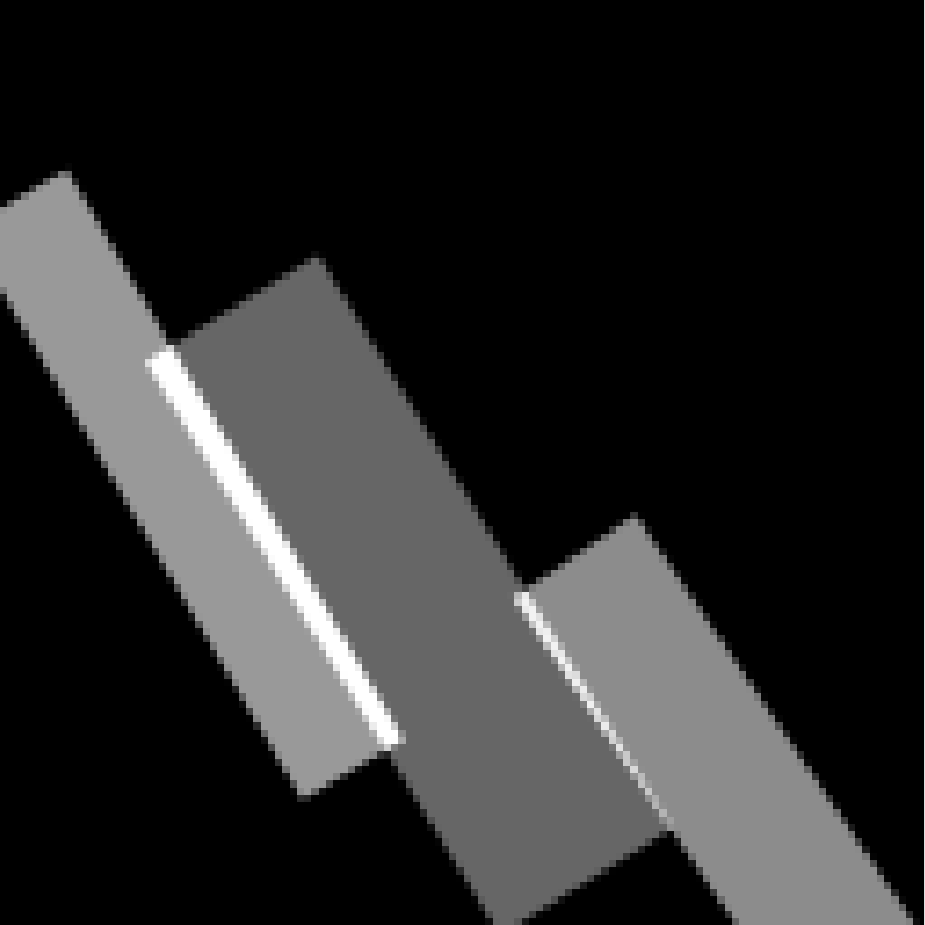}
    \includegraphics[width=0.22\linewidth]{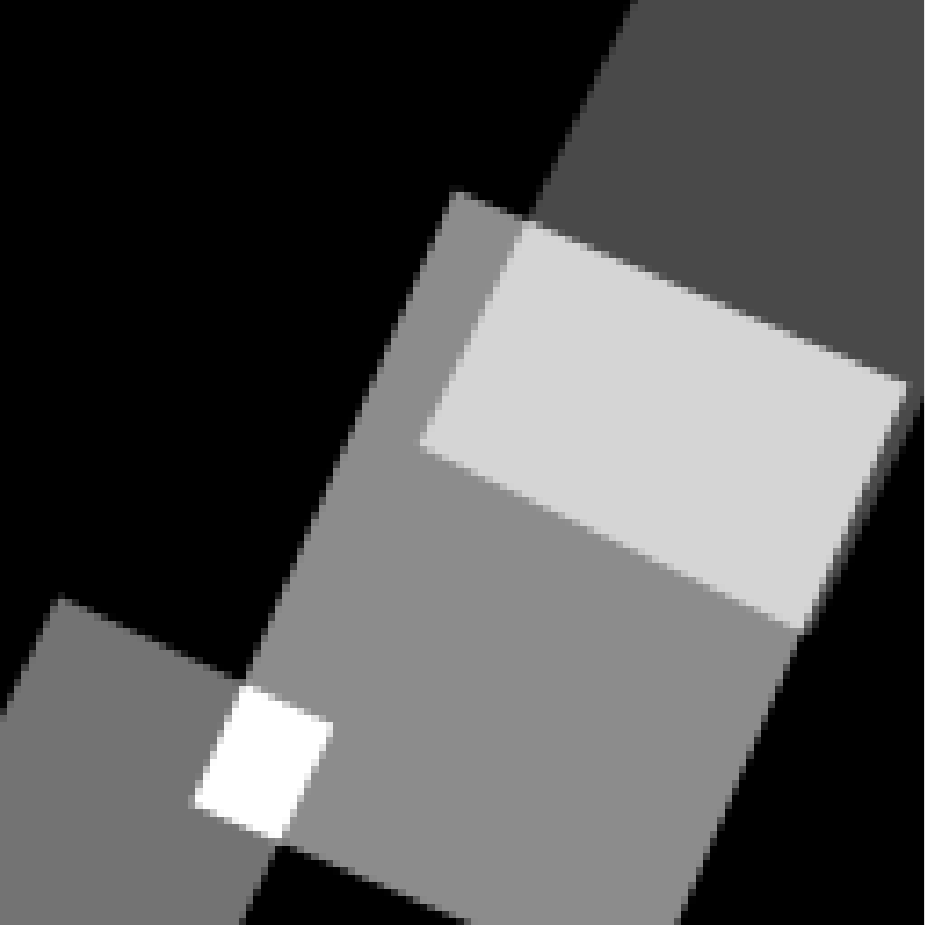}
    \includegraphics[width=0.22\linewidth]{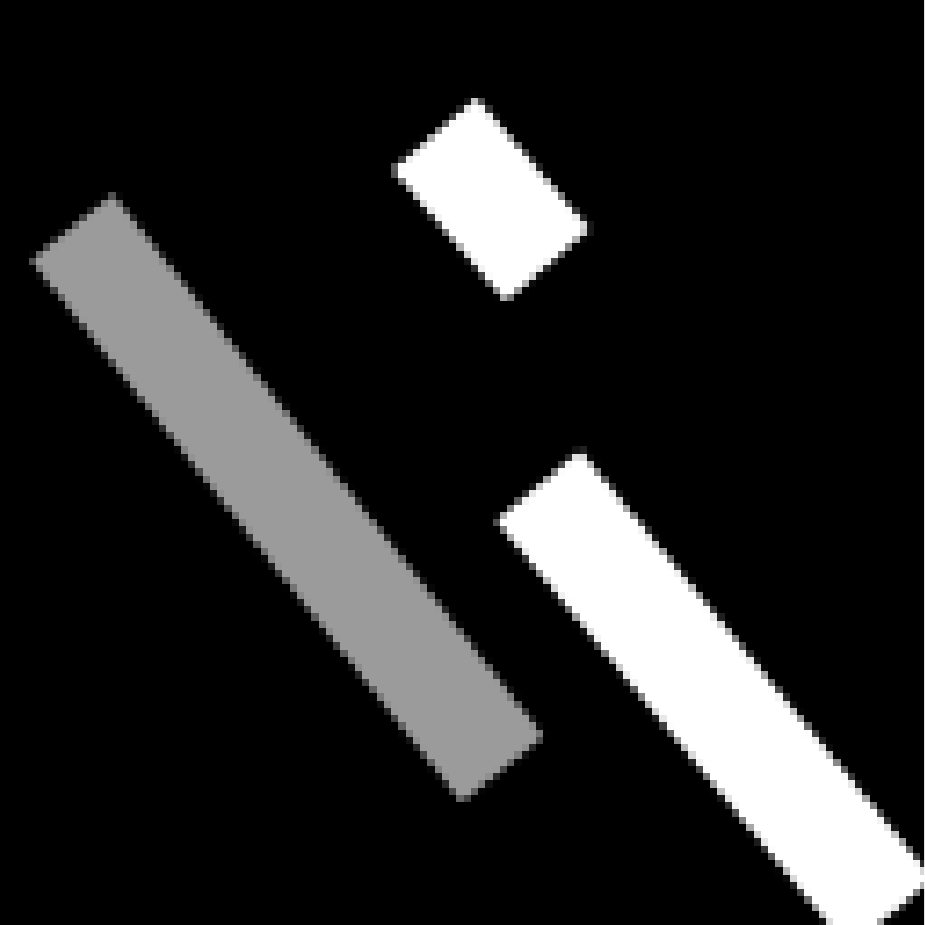}
    \caption{Examples of synthetic images.}
    \label{fig:example_images}
\end{figure}

We use a synthetic dataset comprising images of rectangles with randomised shape, orientation and intensity values, and simulate CT measurements by applying the forward operator $\op \in \BR^{d_{y} \times d_{x}}$ and adding Gaussian noise with standard deviation of 5\,\% or 10\,\% of the average absolute value of the noiseless measurements $\op x$. 
Each image has resolution $128\times 128$ px$^2$ and shows 3 superimposed rectangles, whose orientation is sampled from a single normal distribution with zero mean and standard deviation $2.86^\circ$. Thus, images in this class contain edges in roughly two perpendicular directions.
\Cref{fig:example_images} shows example images from the dataset.

\subsection{Implementation details for the linearised DIP}\label{appx:linearised_DIP}

The key step of efficiently implementing the linearised DIP is the computation and Cholesky decomposition of the measurement covariance matrix $\Sigma_{yy}$. We describe this step in the following paragraphs and refer to \cite{Antoran2022Tomography}, which we have followed in our implementation, for a complete set of details. 

\paragraph{Computing the measurement covariance matrix $\Sigma_{yy}^{(t)}$}

To assemble or multiply with $\Sigma_{yy}^{(t)}$, we employ matrix-free methods. Our workhorses are the matrix vector products $v_{x}^{\top}\Sigma_{xx}$ and $v_{y}^{\top}\Sigma_{yy}^{(t)}$ for $v_{x}\in\BR^{d_{x}}$ and $v_{y}\in\BR^{d_{y}}$. 
We efficiently compute these products through successive matrix vector product with the components of either $\Sigma_{xx}$, or $\Sigma_{yy}^{(t)}$, respectively. For instance, 
\begin{equation*}
    v_{y}^{\top}\Sigma_{yy}^{(t)} = v_{y}^{\top}\left(\op^{(t)}\mJ \Sigma_{\theta} \mJ^{\top} \left(\op^{(t)}\right)^{\top} + \sigma_{y}^{2} \mI \right).
\end{equation*}
For any vector $v_{\theta}$ of appropriate size, we compute Jacobian vector products $v_{\theta}^{\top}\mJ^{\top}$ using forward mode automatic differentiation (AD) and $v_{\theta}^{\top}\mJ$ using backward mode AD. For the non g-prior model, we efficiently compute products with $\Sigma_{\theta}$ by exploiting its block diagonal structure.  Since the g-prior covariance matrix is diagonal, computing products with it is straightforward.

\paragraph{Numerically stable sample generation with Matheron's rule \cref{eq:matheron}}

Numerical instabilities can arise during the sample generation with the Matheron's rule due to the inversion of $\Sigma_{yy}$, updated via \cref{eq:covariance_update}.
We resort to a simple regularisation strategy, which consists in adding to $\Sigma_{yy}^{(t)}$ a small diagonal element $\epsilon \mI$, where $\epsilon$ is chosen from 1\% to 10\% of the diagonal mean, similarly to \cite{lee2020finite}.

\subsection{Hyperparameters for TV and DIP reconstruction}

The TV strength (i.e.\ $\lambda)$ used in the DIP optimisation and the TV regularised objective, reported in \cref{tab:tv_hyperparams} and \cref{tab:dip_hyperparams}, are found by grid search on 10 validation images.
The DIP reconstruction quality from some images degrades when using many iterations \cite{baguer2020diptv}, so an early stopping would be beneficial.
For the PSNR evaluations of DIP reconstructions, we iterate for 30\,000 steps and select the maximum PSNR for each image; this resembles the ideal early stopping by using the (in practice unknown) ground truth image, and is done in order to exclude the complexity of the stopping mechanism from our evaluations. For the DIP optimisations used for angle selection (i.e.\ the initial DIP on $\CB^{(0)}$ and the DIPs retrained every 5 angles), the numbers of iterations in \cref{tab:dip_hyperparams} are used, which were found by grid search on 10 validation images.

\begin{table}
\centering
\begin{tabular}{@{\extracolsep{2mm}}rcccccccc}
& \multicolumn{4}{c}{5\,\% noise} & \multicolumn{4}{c}{10\,\% noise}\\\cline{2-5}\cline{6-9}\\[-0.8em]
\#angles: & 5 & 10-15 & 20-30 & 35-40 & 5 & 10-15 & 20-30 & 35-40 \\[0.2em]\cline{1-1}\cline{2-5}\cline{6-9}\\[-0.8em]
TV strength $\lambda$  &  $1e{-}2$ & $3e{-}3$ & $3e{-}3$ & $3e{-}3$  &  $1e{-}2$ & $1e{-}2$ & $1e{-}2$ & $3e{-}3$ \\
iterations  &  60\,000 & 30\,000 & 10\,000 & 10\,000  &  60\,000 & 30\,000 & 10\,000 & 10\,000 \\\cline{1-1}\cline{2-5}\cline{6-9}
\end{tabular}
\caption{Hyperparameters for TV reconstruction. The values for $\lambda$ are found by grid search on 10 validation images using $5$, $10$, $20$ and $40$ angles, and the numbers of iterations are chosen such that convergence is observed.}
\label{tab:tv_hyperparams}
\end{table}

\begin{table}
\centering
\begin{tabular}{@{\extracolsep{2mm}}rcccccccc}
& \multicolumn{4}{c}{5\,\% noise} & \multicolumn{4}{c}{10\,\% noise}\\\cline{2-5}\cline{6-9}\\[-0.8em]
\#angles: & 5 & 10-15 & 20-30 & 35-40 & 5 & 10-15 & 20-30 & 35-40 \\[0.2em]\cline{1-1}\cline{2-5}\cline{6-9}\\[-0.8em]
TV strength $\lambda$  &  $3e{-}3$ & $3e{-}3$ & $3e{-}3$ & $1e{-}3$  &  $1e{-}2$ & $1e{-}2$ & $3e{-}3$ & $3e{-}3$ \\
iterations  &  19\,000 & 9400 & 12\,000 & 13\,000  &  11\,000 & 7500 & 12\,000 & 7100 \\\cline{1-1}\cline{2-5}\cline{6-9}
\end{tabular}
\caption{Hyperparameters for DIP reconstruction (including TV regularisation). The values are found by grid search on 10 validation images using $5$, $10$, $20$ and $40$ angles.}
\label{tab:dip_hyperparams}
\end{table}

\begin{figure}
    \centering
    \includegraphics[width=0.75\textwidth]{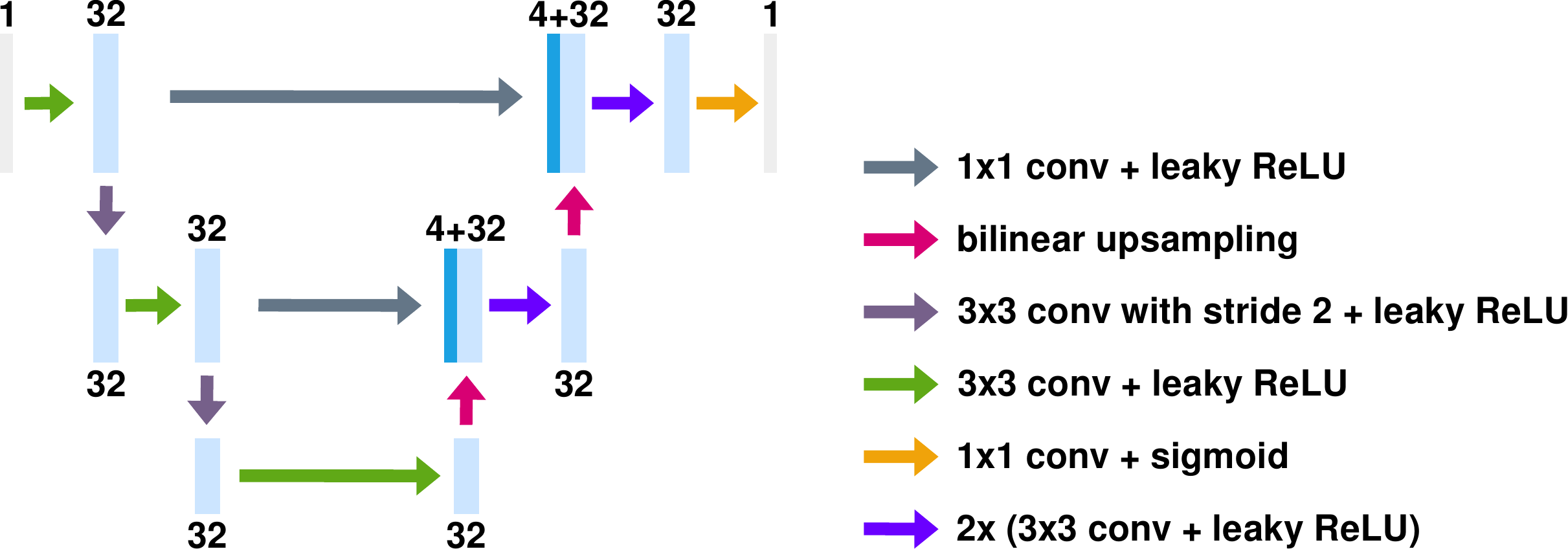}
    \caption{U-net architecture. Each light-blue box corresponds to a
        multi-channel feature map. The number of channels is set to 32
        at every scale. The arrows denote the different operations.
        }
    \label{fig:unet}
\end{figure}

\clearpage

\section{Additional experimental results and analysis}\label{appx:add_exp_results}

In this section, we include additional experimental figures and discuss them.

\begin{figure}[htb]
    \centering
    \includegraphics[width=\linewidth]{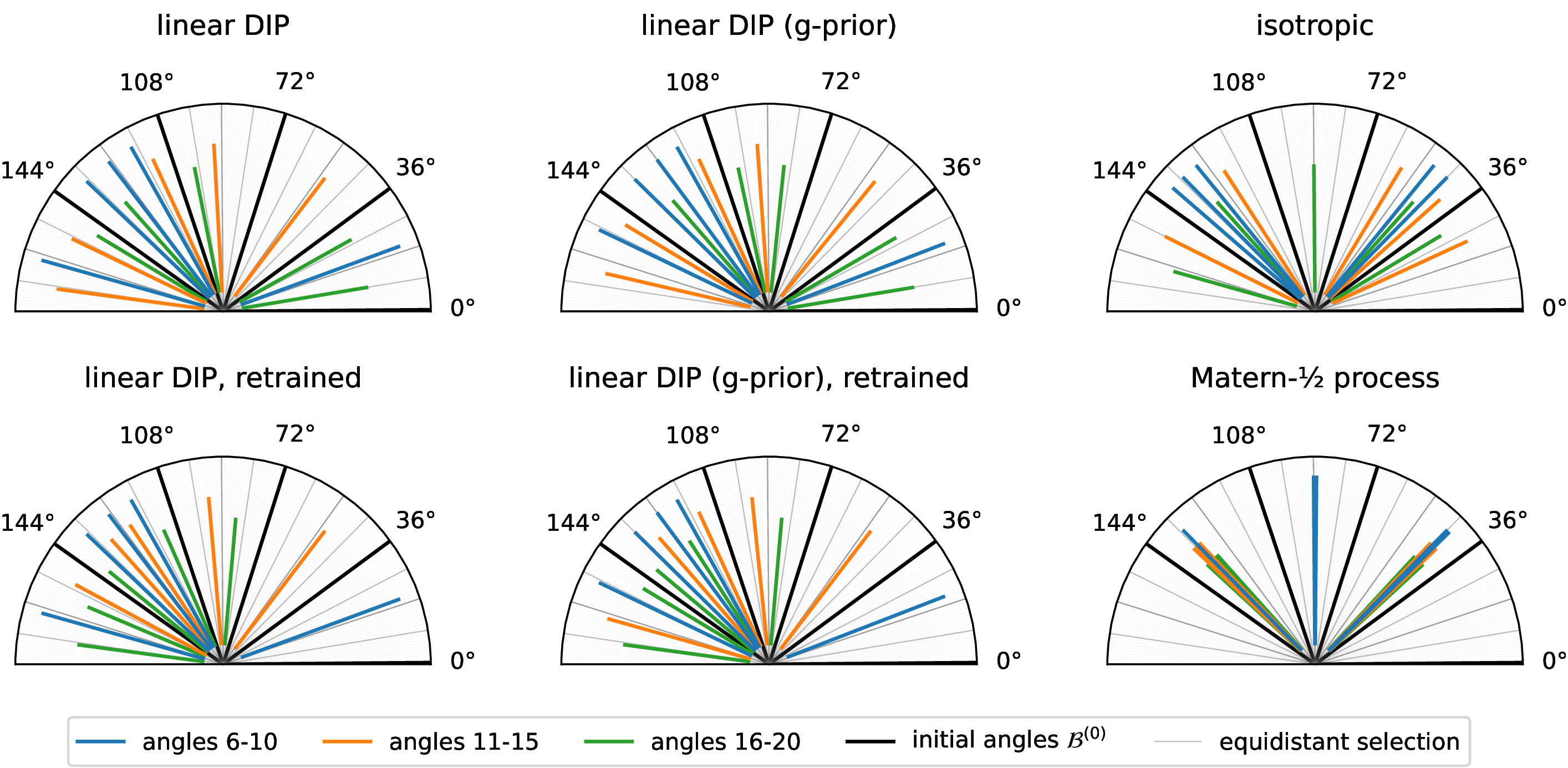}
    \caption{First 20 angles selected by each method under consideration for the example image shown in \cref{fig:variance_angles}}
    \label{fig:angle_selection_image1_complete}
\end{figure}

\Cref{fig:angle_selection_image1_complete} completes \cref{fig:angle_selection_image1} by showing the angles selected by all methods under consideration. Both linear DIP and linear DIP with g-prior choose very similar angles, with the g-prior resulting in a very slightly more diverse angle set. Retraining the linearised DIP every 5 angles to update the basis expansion results in a stronger focus on angles close to the preferential direction. As expected, the differences with the non-retrained DIP are more pronounced for later selected angles (i.e.\ angles 16-20).

The Matern-$\nicefrac{1}{2}$  model concentrates its selection on oblique angles much more strongly than the isotropic model. This results in a very non-diverse angle set which achieves very poor performance. To understand why this happens we first remark that the Matern-$\nicefrac{1}{2}$  model generalises the isotropic model and the two are equal when the lengthscale is set to $\ell=0$. We investigate the hyperparameters chosen by the model evidence for the Matern-$\nicefrac{1}{2}$  model and find that for all images the lengthscale is in the range [40-70]. This value is very large relative to the size of the image ($128 \times 128$) and represents an assumption that the reconstructed image has only 2 or 3 regions with different pixel intensity values. Under this assumption, only taking measurements at 3 different angles is justified.  

\begin{figure}[htb]
\centering
{\sffamily\scriptsize Isotropic model, first 8 acquisitions}\\
\includegraphics[width=\linewidth]{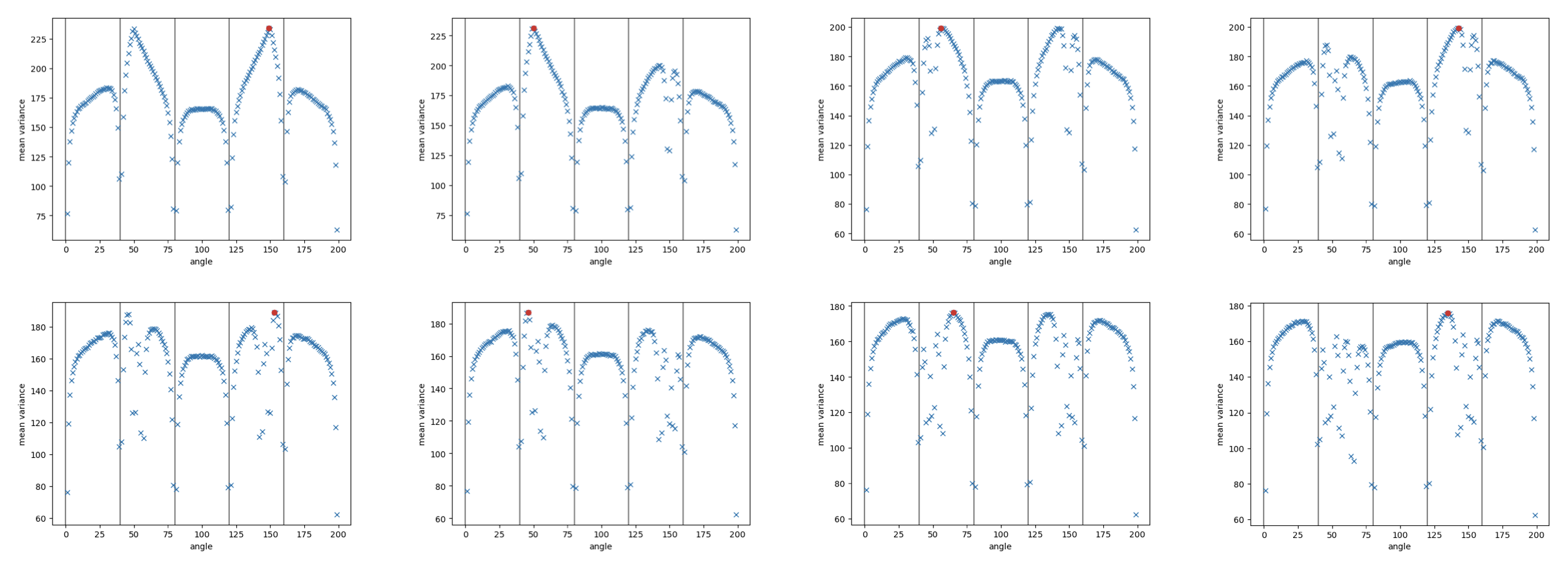}
\caption{Variance assigned to each candidate angle during the first 8 design steps by our Isotropic model.}
\label{fig:isotropic_first_8}
\end{figure}

\begin{figure}[htb]
\centering
{\sffamily\scriptsize Matern-$\nicefrac{1}{2}$ model, first 8 acquisitions}\\
\includegraphics[width=\linewidth]{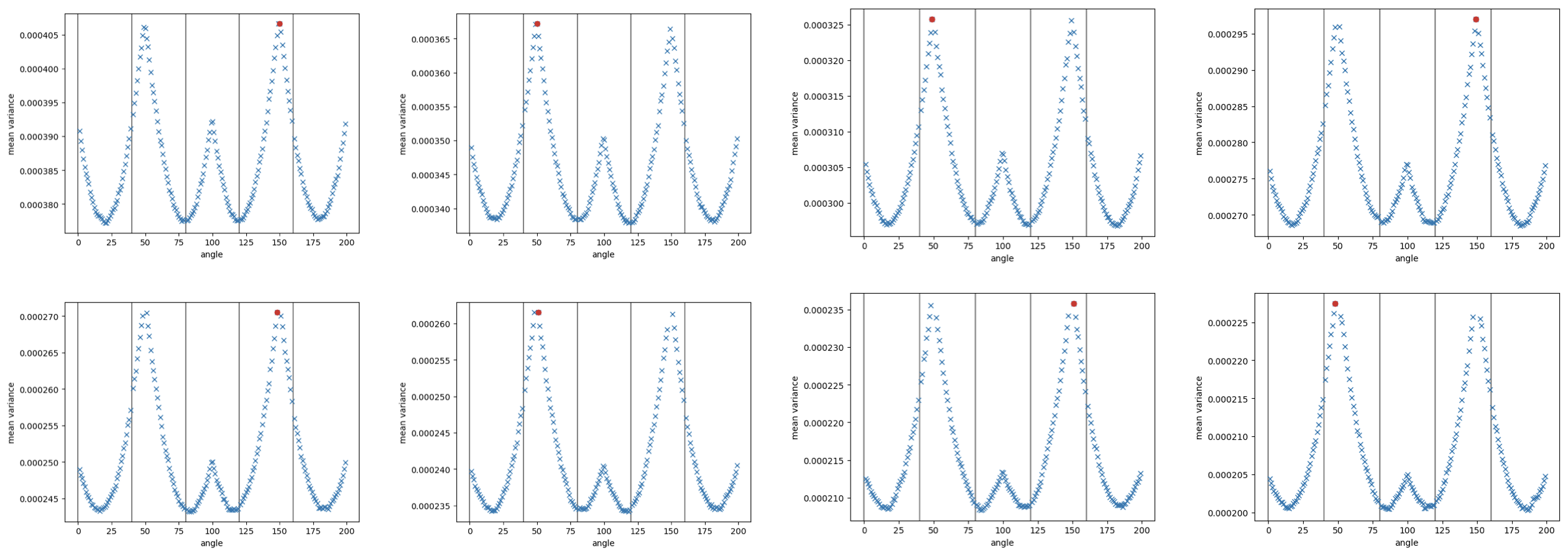}
\caption{Variance assigned to each candidate angle during the first 8 design steps by our Matern-$\nicefrac{1}{2}$  model.}
\label{fig:gp_first_8}
\end{figure}

We verify this explanation by examining the ESE scores assigned by the isotropic and Matern-$\nicefrac{1}{2}$  models to the first 8 angles chosen in \cref{fig:isotropic_first_8} and \cref{fig:gp_first_8} respectively. The isotropic model chooses oblique angles. After each new angle is included in the updated operator $\op^{(t)}$, the predictive variance in a region spanning roughly $10^\circ$ around the chosen angles decreases. This is the span of the detector elements. The uncertainty at other angles remains unchanged because the model assumes reconstruction pixels to be uncorrelated. By modelling correlations among detector pixels, each additional angle should reduce the Matern-$\nicefrac{1}{2}$  model's uncertainty in a larger angle range (set via the lengthscale), promoting exploration. However, because the lengthcale, which has overfit the pilot measurements, is very large, each new angle introduced into the operator reduces the predictive variance of every angle almost equally. 
As a result, the relative assignment of predictive variance in angle space remains roughly constant throughout design steps, and all of the chosen angles become very similar to each other.

Although, it is well known that experimental design is very sensitive to the choice of prior \citep{chi2015modelerror,foster2021Design}, the ease with which the relatively very simple Matern-$\nicefrac{1}{2}$  model can overfit the degree to which this degrades performance was unexpected to us.
In future work we will investigate alternative methods for setting model hyperparameters.

\begin{figure}[htb]
\centering
{\sffamily\scriptsize Linearised DIP, first 8 acquisitions}\\
\includegraphics[width=\linewidth]{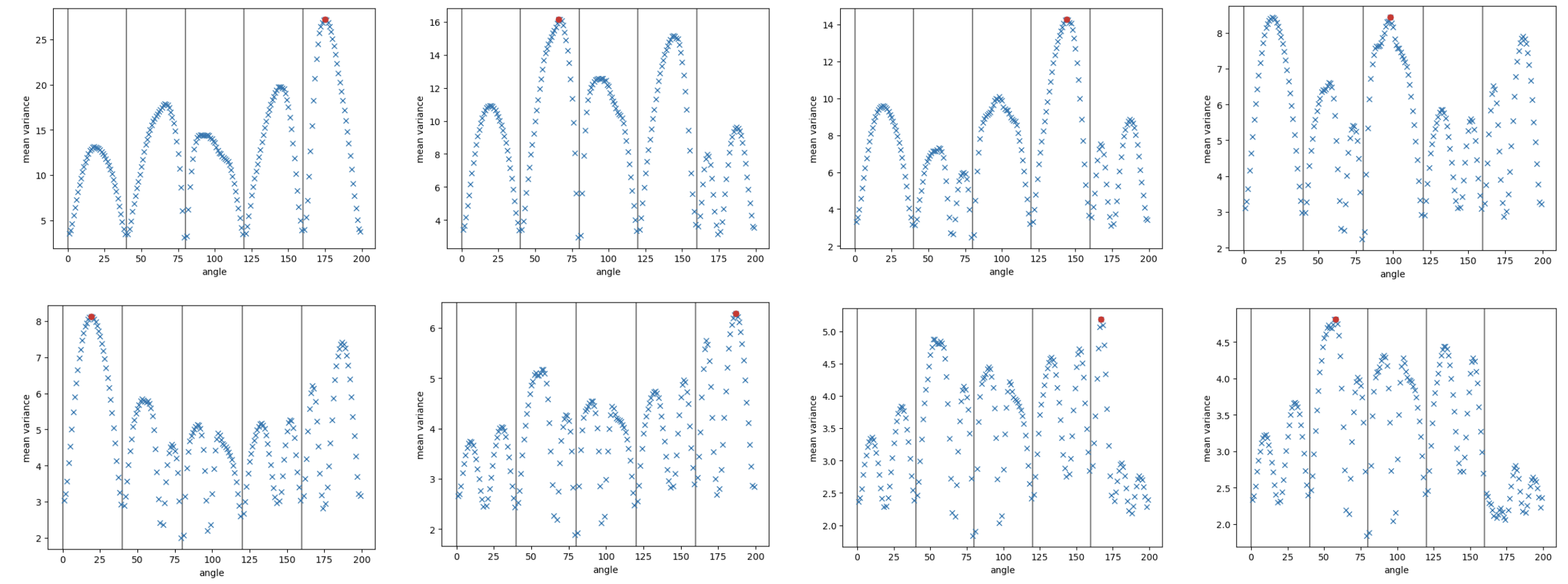}
\caption{Variance assigned to each candidate angle during the first 8 design steps by our linearised DIP model.}
\label{fig:mll_first_8}
\end{figure}

\begin{figure}[htb]
\centering
{\sffamily\scriptsize Linearised DIP with g-prior, first 8 acquisitions}\\
\includegraphics[width=\linewidth]{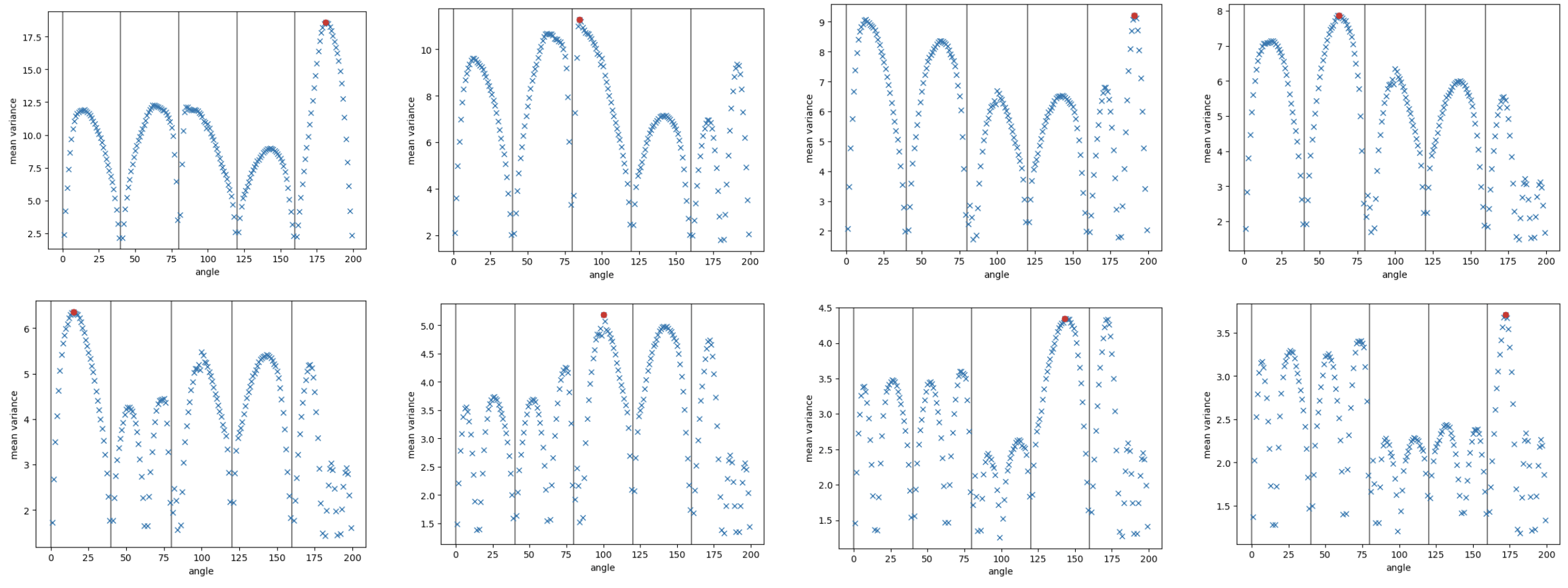}
\caption{Variance assigned to each candidate angle during the first 8 design steps by our linearised DIP model with the g-prior.}
\label{fig:g-prior_first_8}
\end{figure}

\Cref{fig:mll_first_8} and \cref{fig:g-prior_first_8} show the variance assigned to each angle in the first 8 acquidition steps on an example image (first image from \cref{fig:example_images}) for the linearised DIP and the linearised DIP with g-prior, respectively. Although the angles selected by the two models are different, both prioritise similar angle regions.

\begin{figure}[!htb]
\minipage{0.32\textwidth}
  \includegraphics[width=\linewidth]{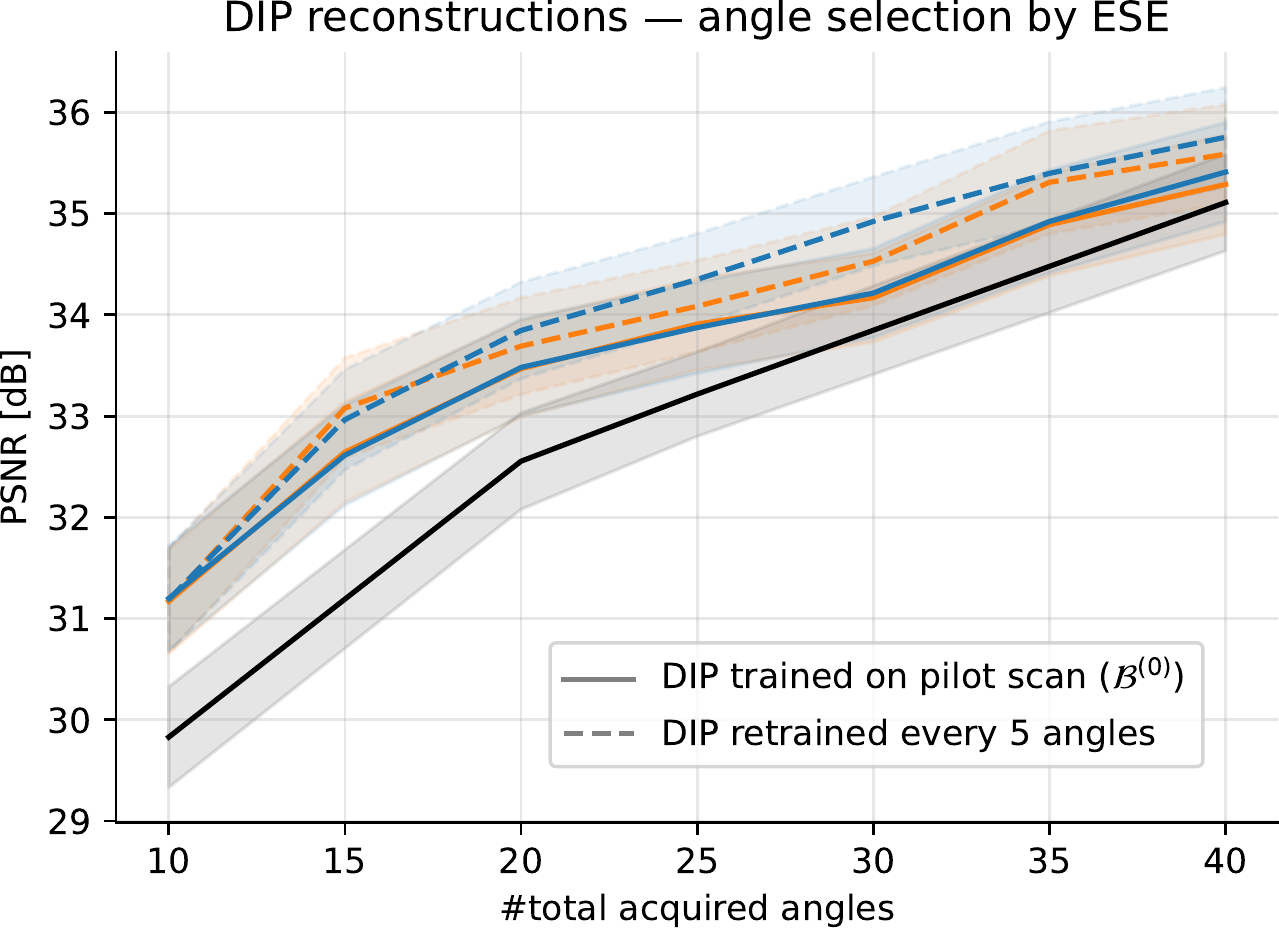}
\endminipage\hfill
\minipage{0.32\textwidth}
  \includegraphics[width=\linewidth]{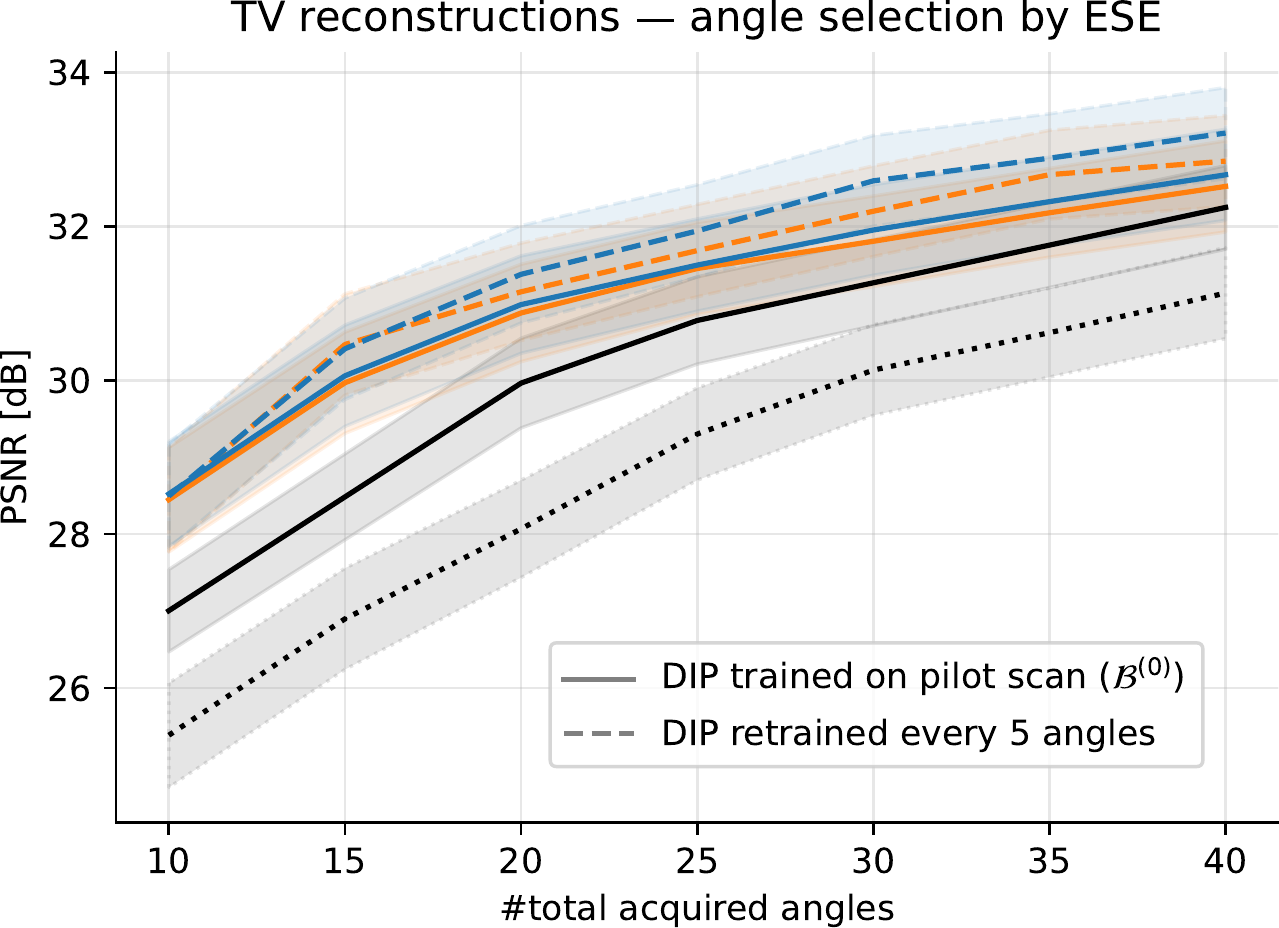}
\endminipage\hfill
\minipage{0.32\textwidth}%
  \includegraphics[width=\linewidth]{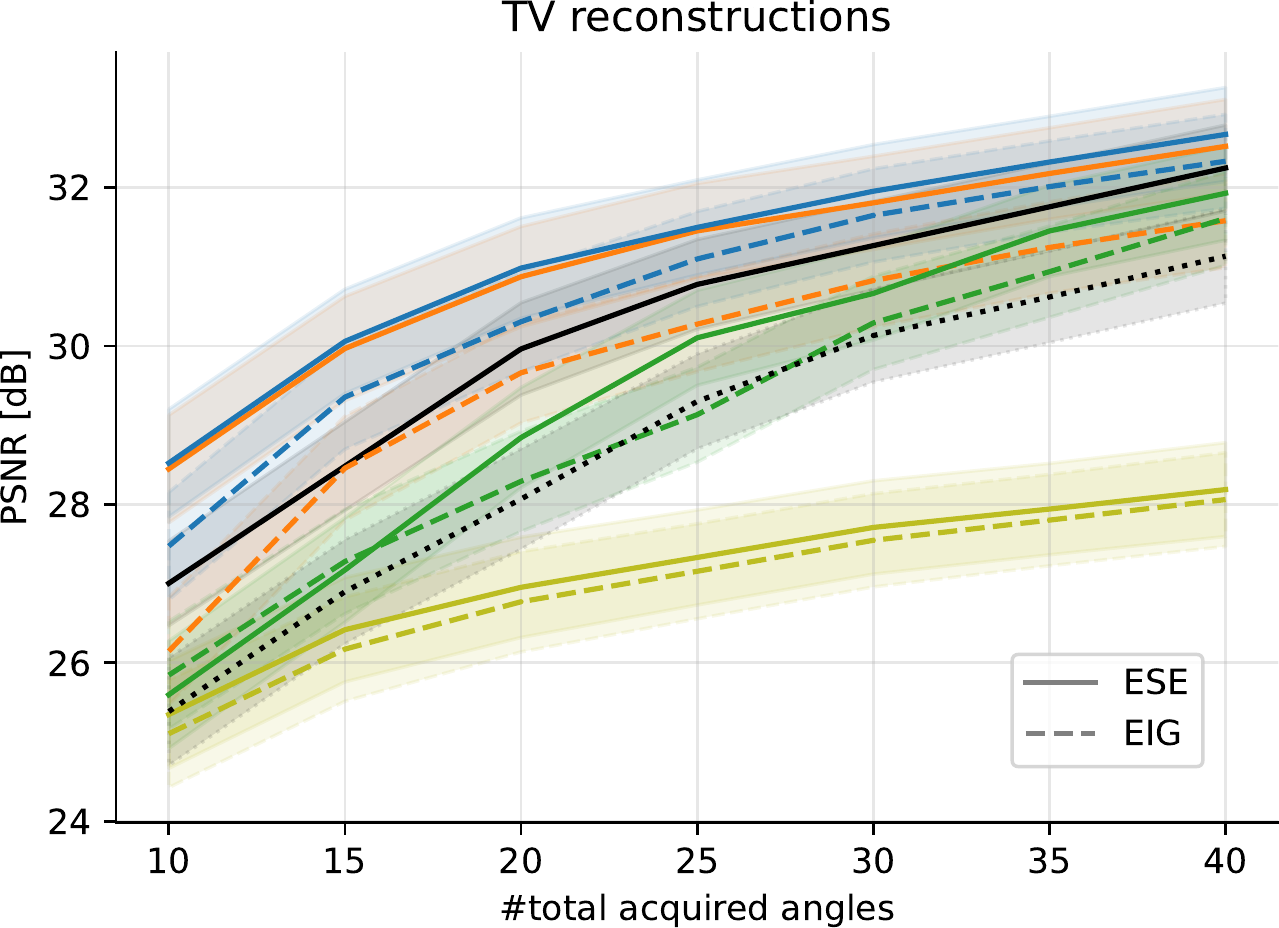}
\endminipage
\\[1em]
\centering
\includegraphics[width=.85\linewidth]{images/main_figure/legend.pdf}
\caption{Reconstruction PSNR vs n. angles scanned, averaged across 30 images (5\% noise).}\label{fig:main-figure-psnr-comparison-std-included}
\end{figure}

\Cref{fig:main-figure-psnr-comparison-std-included} is a more complete version of \cref{fig:main-figure-psnr-comparison}, including the standard error. Given our 30 image runs, we can conclude that the linearised DIP provides a statistically significant improvement over the equidistant baseline up to 20 selected angles. By retraining the DIP Jacobians every 5 angles, we can extend the significant improvements up to 35 scanned angles. In future we aim to make these statements stronger by running more experiments.

\begin{figure}[!htb]
\minipage{0.32\textwidth}
  \includegraphics[width=\linewidth]{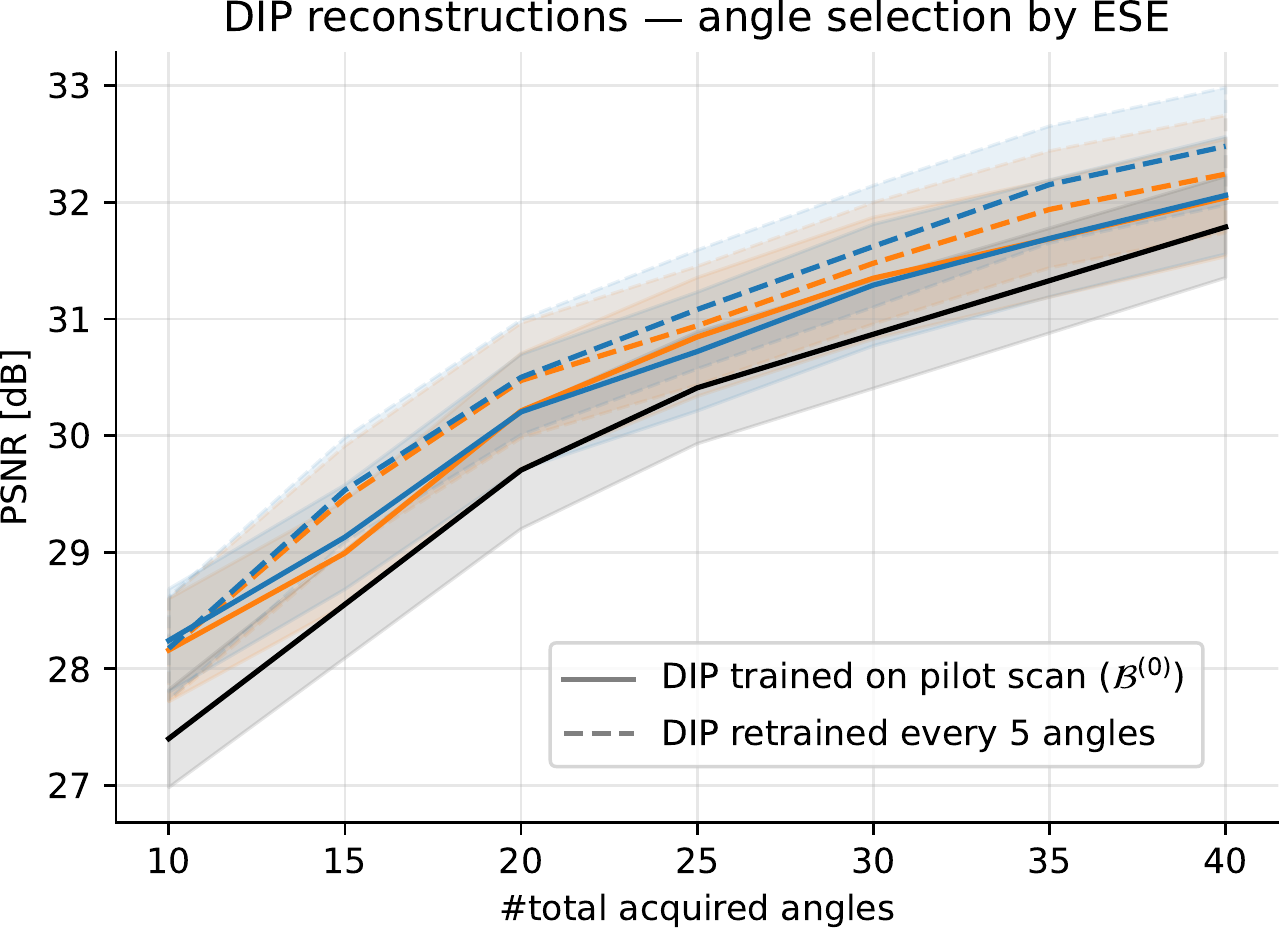}
\endminipage\hfill
\minipage{0.32\textwidth}
  \includegraphics[width=\linewidth]{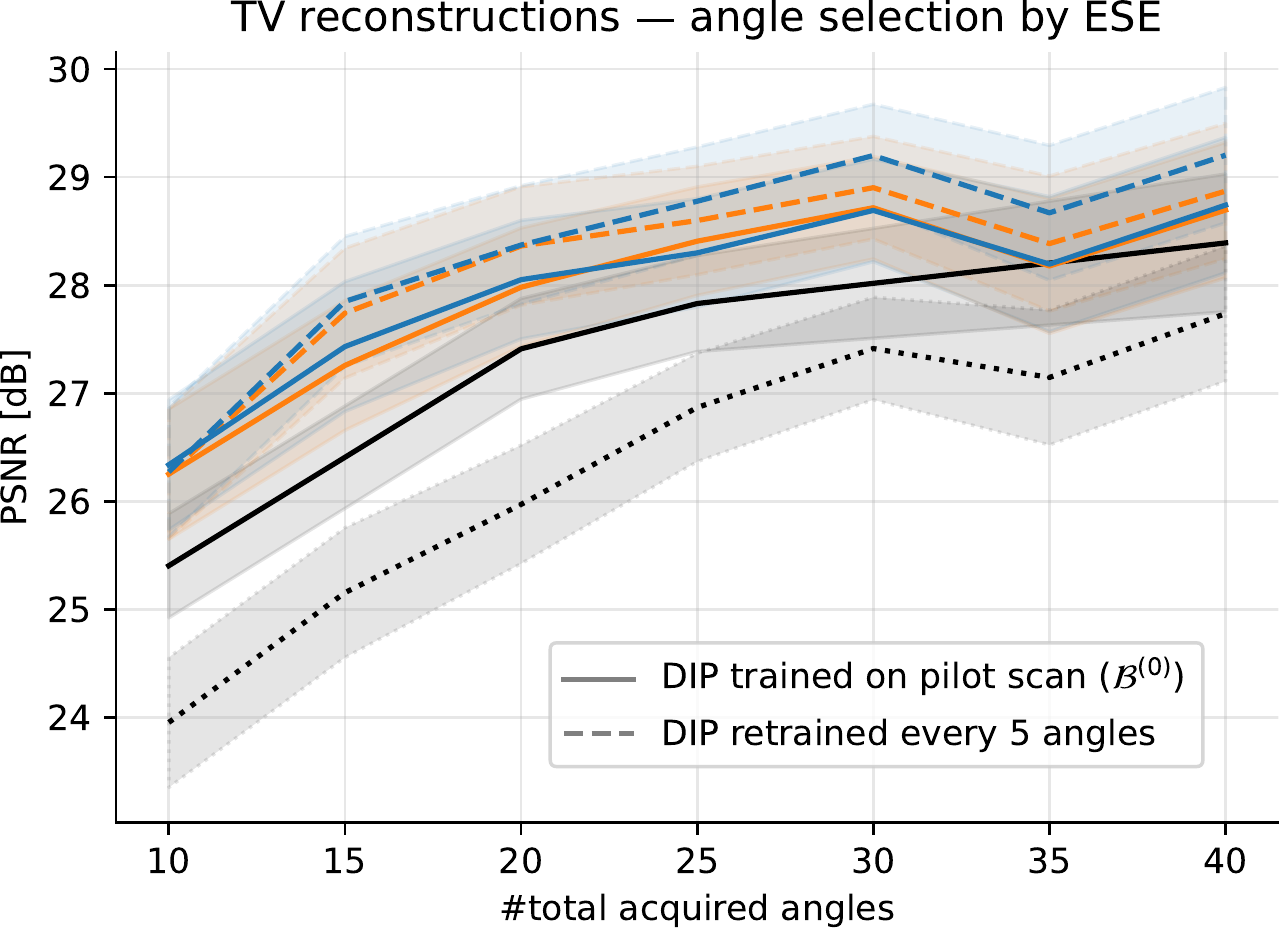}
\endminipage\hfill
\minipage{0.32\textwidth}%
  \includegraphics[width=\linewidth]{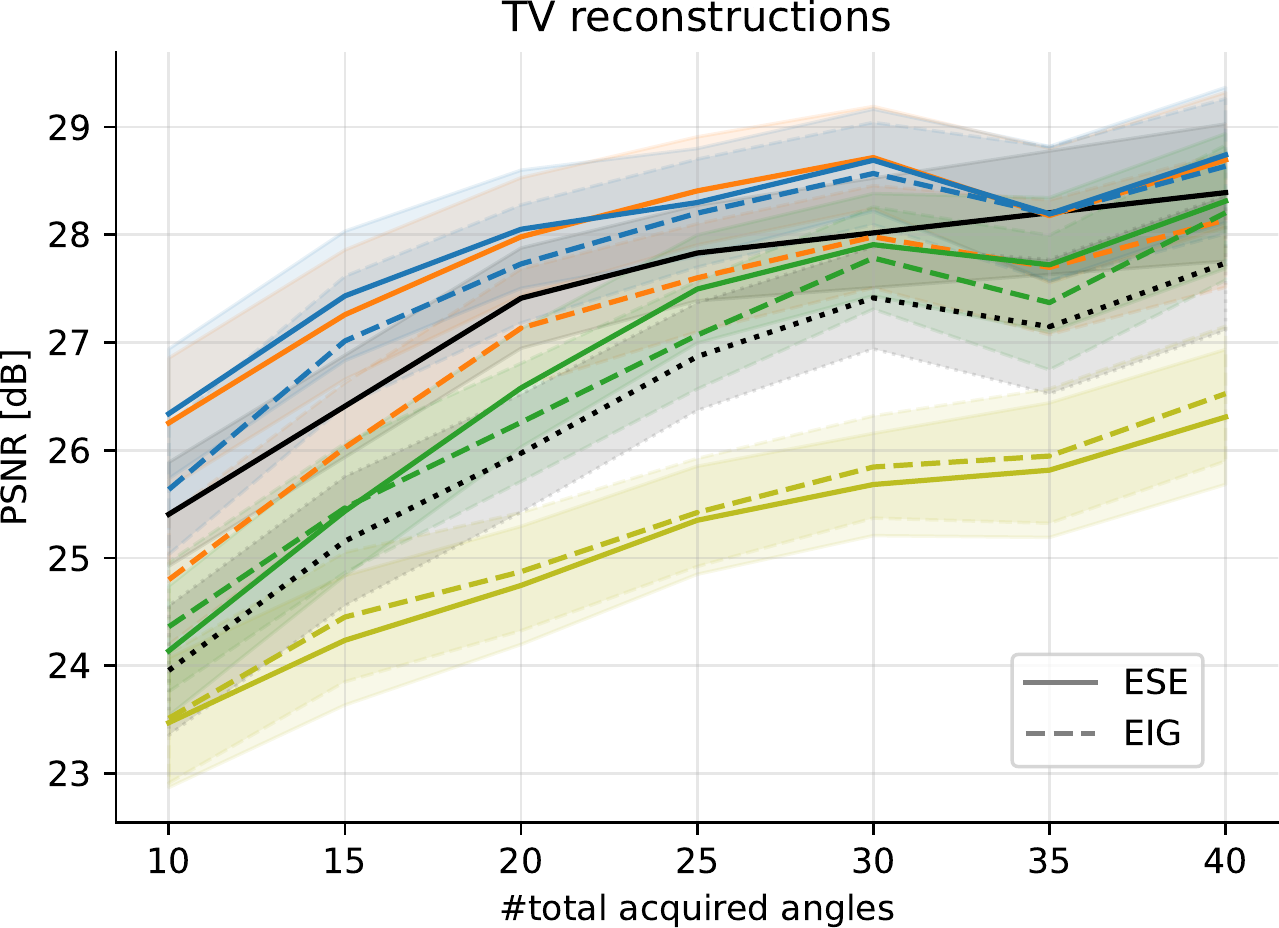}
\endminipage
\\[1em]
\centering
\includegraphics[width=.85\linewidth]{images/main_figure/legend.pdf}
\caption{Reconstruction PSNR vs n. angles scanned, averaged across 30 images (10\% noise).}\label{fig:main-figure-psnr-comparison-std-included-01} 
\end{figure}

\Cref{fig:main-figure-psnr-comparison-std-included-01} shows our findings on measurement data simulated adding $10\%$ noise. The gains from experimental design are slightly reduced in the noisier setting, although the conclusions remain the same. From the EIG expression \cref{eq:eig}, we can see that noisier measurements should push our score assignment to be more uniform across angles and thus closer to the equidistant baseline. 

\end{document}